\documentclass[twoside,11pt]{article}

\usepackage[preprint]{jmlr2e}

\usepackage{amsmath}
\usepackage{amssymb}
\usepackage{hyperref}

\usepackage{multirow}
\usepackage{tocloft}

\usepackage{graphicx}
\usepackage{booktabs}
\usepackage{wrapfig}
\usepackage{subcaption}

\usepackage{tikz}
\usetikzlibrary{positioning,arrows.meta}
\definecolor{steelblue}{RGB}{70,130,180}

\ShortHeadings{FlowGenesis}{Kazanskii}

\title{FlowGenesis: Representation-Driven Self-Organization
of Neural Architectures}

\author{
  \name Maksim Kazanskii
  \email mkazanskii@gmail.com \\
  \addr Independent Researcher
}

\begin{document}

\maketitle

\begin{abstract}
Deep neural networks typically rely on architectures specified before training, despite the continuous evolution of their internal representations. We introduce FlowGenesis, a framework in which neural architectures adapt through neuron birth and pruning guided by Representation Flux, a local statistic capturing changes in minibatch mean activations. Neurons exhibiting persistently high Flux undergo splitting designed to approximately preserve the network function, while neurons with low Flux and low activity are pruned. This mechanism allows network capacity to evolve without prescribing a target architecture or performing a separate architecture search. Experiments on image and text classification show that networks initialized with different sizes approach similar final neuron-count ranges within each dataset while maintaining comparable predictive performance. We further extend FlowGenesis to continual learning, where representation-guided structural adaptation reduces the neurons and parameters allocated to successive tasks while maintaining accuracy comparable to fixed-width Progressive Neural Networks in a task-incremental learning setting. These findings demonstrate that neural architectures can adapt through feedback between representation learning and structural change, suggesting an alternative to prescribing network capacity before training.
\end{abstract}

\begin{keywords}
neurogenesis, structural plasticity,
representation flux, architectural attractors,
self-organization, continual learning
\end{keywords}

\section{Introduction}

Deep neural networks have achieved remarkable success across machine
learning tasks, yet their architectures are typically specified before
optimization. Adaptive structural learning, network growth and pruning,
dynamic sparsity, neuron splitting, and developmental architectures
challenge this paradigm
\citep{elsken2019nas,cortes2017adanet,wu2019splitting,
evci2022gradmax,maile2022neurons,najarro2023ndp}. Existing approaches guide structural adaptation using gradients,
activation or weight geometry, parameter importance, task-level
signals, or learned developmental rules. Temporally accumulated
internal statistics have also been used to guide structural
plasticity \citep{jia2026smgrnn}. However, the temporal evolution
of learned representations remains comparatively unexplored as
a direct signal for architectural development. Representations evolve during optimization, sometimes substantially
even when predictive performance remains stable
\citep{pashakhanloo2023drift}. Related findings on loss of plasticity during continued
optimization further motivate investigating adaptive
computational capacity \citep{lyle2022capacity}.

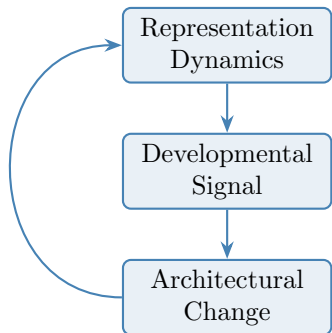
\begin{wrapfigure}{l}{0.38\textwidth}
\centering
\vspace{-0.8em}

\begin{tikzpicture}[
    node distance=0.65cm,
    block/.style={
        rectangle,
        rounded corners=3pt,
        draw=steelblue,
        fill=steelblue!12,
        thick,
        align=center,
        text width=2.5cm,
        minimum height=0.7cm,
        font=\small
    },
    arrow/.style={
        ->,
        >=Stealth,
        thick,
        draw=steelblue
    }
]

\node[block] (rep) {Representation\\Dynamics};
\node[block, below=of rep] (signal) {Developmental\\Signal};
\node[block, below=of signal] (arch) {Architectural\\Change};

\draw[arrow] (rep) -- (signal);
\draw[arrow] (signal) -- (arch);
\draw[arrow]
    (arch.west)
    to[out=180,in=180,looseness=1.5]
    (rep.west);

\end{tikzpicture}

\caption{Feedback between representation dynamics
and architectural adaptation in FlowGenesis.}
\label{fig:neurogenesis_feedback}

\vspace{-0.7em}
\end{wrapfigure}

We therefore ask whether persistent changes in neuronal activation statistics can serve as indicators of demand for additional capacity and guide architectural growth and pruning. To investigate this question, we introduce \textbf{FlowGenesis},
a framework that uses Representation Flux, based on differences in neuronal activation statistics across consecutive minibatches, to regulate network structure.
Neurons exhibiting sustained high Flux undergo
Flux-Driven Neurogenesis, splitting into units that can
subsequently specialize. Conversely, neurons with persistently
low Flux and low activity become candidates for pruning.
Requiring both conditions reduces the risk of treating stable but active representations as redundant. These mechanisms establish a feedback loop between
representation dynamics and architecture, as illustrated
in Figure~\ref{fig:neurogenesis_feedback}.

This mechanism motivates investigating whether Flux-guided neuron selection influences architectural development and predictive performance relative to alternative selection strategies. A second question concerns the emergence of reproducible, task-dependent architectural regimes despite differences in initial network capacity. We investigate whether FlowGenesis exhibits such attractor-like behavior across image and text classification tasks. 

Continual learning provides a complementary setting
for investigating adaptive capacity allocation across
sequential tasks. We extend FlowGenesis to a progressive
task-incremental architecture, where previously learned
modules remain frozen and new task-specific modules
adapt their capacity through neuron birth and pruning.

To be precise, we investigate three research questions:

\begin{enumerate}
    \item How does Flux-guided neuron selection affect
    architectural development and predictive performance
    compared with alternative allocation strategies?

    \item Do different initial architectures approach reproducible, task-dependent neuron-count regimes?

    \item Can adaptive capacity allocation reduce network
    growth while maintaining predictive performance
    in task-incremental continual learning?
\end{enumerate}

\section{Related Work}
\label{sec:related_work}

FlowGenesis lies at the intersection of neural architecture
search, constructive neural networks, developmental and
self-organizing architectures, and representation dynamics.
We review these directions with particular attention to
the signals governing structural adaptation.

\subsection{Neural Architecture Search and Adaptive Architectures}

Neural architecture search (NAS) optimizes network structure over
a predefined search space, using methods ranging from reinforcement
learning and evolutionary optimization to differentiable approaches
such as DARTS \citep{elsken2019nas,liu2019darts}. AdaNet constructs
networks adaptively while balancing predictive performance and model
complexity \citep{cortes2017adanet}. Function-preserving approaches,
including Net2Net and Network Morphism, enable architectural expansion
or transformation while preserving the function represented by the
network \citep{chen2016net2net,wei2016networkmorphism}.

\subsection{Network Growth, Neurogenesis, and Pruning}

Constructive neural networks adapt computational capacity during
training. Splitting Steepest Descent (SSD) grows networks by splitting
neurons according to a second-order functional descent criterion
\citep{wu2019splitting}. Firefly Neural Architecture Descent extends
this perspective to progressive architectural growth
\citep{wu2020firefly}. GradMax introduces neurons using gradient
information, selecting initializations that facilitate subsequent
optimization \citep{evci2022gradmax}. NeST combines growth and
pruning within a unified synthesis procedure \citep{dai2019nest},
while dynamic sparse training methods such as RigL modify network
connectivity during optimization under a largely fixed parameter
budget \citep{evci2020rigl}.

Particularly relevant is the framework of \citet{maile2022neurons},
which formulates artificial neurogenesis in terms of \emph{when},
\emph{where}, and \emph{how} neurons should be added. Their NORTH*
methods use activation or weight orthogonality to guide capacity
expansion, demonstrating that internal geometric statistics can
inform structural adaptation.

Together, these methods establish several criteria for allocating
capacity, including loss geometry, gradients, activation geometry,
and parameter structure. They also distinguish the decision to
introduce capacity from the mechanism by which new computational
units are initialized and integrated into an existing network.

\subsection{Developmental and Self-Organizing Neural Networks}

Developmental approaches allow network structure to emerge through
local learning and adaptation. Self-Building Neural Networks combine
Hebbian learning with pruning to adapt structure during training
\citep{ferigo2023selfbuilding}. Neural Developmental Programs (NDPs)
generate architectures through learned local developmental rules
\citep{najarro2023ndp}, while Lifelong NDPs extend this approach
to structural and synaptic plasticity in non-stationary environments
\citep{plantec2024lndp}.

Particularly relevant are Self-Motivated Growing Neural Networks
(SMGrNN), which use neuronal activity and temporally accumulated
weight-update statistics to trigger neuron insertion and pruning
\citep{jia2026smgrnn}. This demonstrates that online structural
adaptation can be governed by internal signals accumulated over
training, without conventional architecture search.

In contrast to SMGrNN, which relies on accumulated
weight-update statistics, FlowGenesis uses temporal
differences in neuronal activation statistics to
guide neuron birth and pruning. Unlike developmental
approaches based on learned local rules, FlowGenesis
employs explicit, predefined criteria based on
Representation Flux and neuronal activity.

\subsection{Structural Plasticity and Continual Learning}

Structural adaptation provides a mechanism for accommodating
sequentially changing learning problems. Dynamically Expandable
Networks allocate additional capacity as new tasks arrive,
selectively retraining and expanding the existing network
\citep{yoon2018den}. More broadly, dynamic expansion and
parameter-isolation methods mitigate interference by allocating
or protecting different subsets of parameters across tasks.
Progressive Neural Networks preserve previously learned
task-specific parameters while introducing new network
columns for subsequent tasks \citep{rusu2016progressive}.

An alternative approach maintains plasticity through continuous
feature renewal. Continual Backprop employs a generate-and-test
mechanism that replaces low-utility features with newly
initialized ones, preserving adaptability during prolonged
learning \citep{dohare2021continual}. Unlike task-dependent
expansion, this mechanism does not require explicit task
boundaries to initiate feature replacement.

FlowGenesis differs from these approaches in how capacity
is allocated. Unlike task-level expansion methods, which
introduce additional capacity according to task requirements,
FlowGenesis regulates neuron birth and pruning within
task-specific modules using Representation Flux and activity
statistics. In the progressive continual-learning setting,
previously learned modules remain frozen, while new modules
adapt their capacity during training. Thus, FlowGenesis
combines parameter isolation with representation-guided
structural adaptation, rather than relying on fixed-width
expansion or utility-based feature replacement.

\subsection{Connection to Neuroscience}

Biological neural systems exhibit structural plasticity
through activity-dependent synaptic remodeling, developmental
pruning, and neurogenesis in restricted adult neural
populations \citep{holtmaat2009experience,
paolicelli2011synaptic,ming2011adult}.

Adult neurogenesis provides an example of activity-dependent
capacity addition. In the mammalian hippocampus, newly
generated neurons integrate into existing circuits, while
their survival and functional integration are influenced
by neuronal activity and synaptic input
\citep{ming2011adult,tashiro2006nmda}.
This process illustrates how the addition and retention
of new computational elements can depend on local
circuit conditions.

Conversely, developmental synaptic pruning selectively
eliminates connections during circuit maturation.
Microglia contribute to this process, with experimental
evidence showing that synaptic elimination depends
partly on neuronal activity and complement signaling
\citep{paolicelli2011synaptic,schafer2012microglia}.

These biological processes provide conceptual inspiration
for FlowGenesis: network structure is not fixed, but can
be modified through activity-dependent addition and
removal. However, biological neurogenesis generates new
neurons from precursor cells, whereas FlowGenesis splits
existing artificial neurons. Likewise, biological synaptic
pruning removes connections, while FlowGenesis can remove
entire computational units. The analogy therefore concerns
adaptive structural regulation rather than a shared
biological mechanism.

\subsection{Architectural Convergence and Emergent Organization}

Self-organizing neural systems demonstrate how global
organization can emerge from local adaptation rules.
Neural gas and growing neural gas algorithms organize
prototype networks according to the distribution of
observed data \citep{martinetz1993neural,
fritzke1995growing}. More broadly, adaptive network
models demonstrate how local connectivity rules can
produce complex global organization
\citep{gross2008adaptive}.

Unlike these approaches, FlowGenesis regulates neuron
birth and pruning using temporal changes in activation
statistics during gradient-based learning. It therefore
investigates whether representation-guided structural
adaptation can produce reproducible architectural
configurations across different initializations.

\subsection{Representation Dynamics and Representational Drift}

Representational drift describes changes in learned
representations over time, even when predictive performance
remains stable. \citet{pashakhanloo2023drift} demonstrate
SGD-induced diffusion-like drift along a minimum-loss
manifold in a two-layer linear network. Related work shows that networks can progressively lose
the ability to learn changing prediction targets
\citep{lyle2022capacity}. 

More directly,
\citet{kazanskii2026geometry} associates elevated
Representation Flux with catastrophic forgetting,
including cases where increased Flux precedes
performance degradation.

These findings motivate using representation dynamics
to guide structural adaptation. Unlike representational
drift, FlowGenesis Flux measures differences in neuronal
activation statistics across consecutive minibatches,
capturing both optimization-induced changes and sampling
variability. FlowGenesis investigates its utility as
a local signal for neuron birth and pruning.

\subsection{Homeostatic Structural Plasticity}

Homeostatic plasticity regulates neuronal activity
in response to sustained changes in network dynamics,
helping maintain stable activity regimes during
learning and development
\citep{turrigiano2004homeostatic,turrigiano2012homeostatic}.

Structural homeostasis extends this principle to
network connectivity. Computational models show
how activity-dependent formation and elimination
of synaptic connections can reorganize network
structure while regulating neuronal activity
\citep{butz2013structural,butz2014structural}.

FlowGenesis shares the principle of locally regulated
structural adaptation but differs from homeostatic
models: Representation Flux measures temporal
differences in activation statistics rather than
deviations from an activity set point. Whether
Flux-guided neuron birth and pruning produce
stable architectural regimes is therefore an
empirical question rather than an assumed
homeostatic mechanism.

\section{Emergent Architectural Development}
\label{sec:emergent_development}

We study architectural development in a minimal neural system consisting of a sparse recurrent population. Unlike highly structured architectures such as convolutional networks or Transformers, this setting allows direct control over architectural properties and observation of individual neuron development. Our objective is not to maximize predictive performance, but to investigate whether local representational dynamics can guide the self-organization of network capacity during learning.

\subsection{Random Neural Container}
\label{sec:container_setting}

The Random Neural Container is based on a sparsely
connected recurrent neural network with leaky-integrator
dynamics. Its recurrent connectivity is randomly
initialized, while the network weights are optimized
through gradient-based training. Unlike classical
reservoir computing \citep{jaeger2001echo,maass2002realtime},
the recurrent weights are trainable, and FlowGenesis
additionally modifies the network topology through
neuron birth and pruning.

Given a fixed input projection $u=W_{\mathrm{in}}x$
and initial state $h^{(0)}=0$, the network performs
$S$ recurrent relaxation steps:
\begin{equation}
h^{(s+1)}=(1-\gamma)h^{(s)}
+\gamma\operatorname{ReLU}
\left(u+W_{\mathrm{eff}}h^{(s)}+b\right),
\end{equation}
where $\gamma=0.2$ is the relaxation coefficient
and recurrent parameters are shared across steps.
The final state is mapped to classification logits:
\begin{equation}
f(x)=W_{\mathrm{out}}h^{(S)}+b_{\mathrm{out}}.
\end{equation}

Recurrent connectivity is represented by a binary mask
$M\in\{0,1\}^{N\times N}$, initially sampled
independently with probability
$\Pr(M_{ij}=1)=\beta$.
Each recurrent weight is normalized by the number
of incoming connections to its target neuron:
\begin{equation}
(W_{\mathrm{eff}})_{ij}
=
\frac{W_{ij}M_{ij}}
{\max(1,\sum_k M_{ik})}.
\end{equation}
The connectivity mask subsequently evolves through
structural adaptation.

The initial architecture is characterized by neuron
count $N_0$ and connectivity probability $\beta$,
controlling population size and recurrent sparsity,
respectively.

\subsection{Representation Flux}
\label{sec:representation_flux}

Structural adaptation is guided by temporal changes in
neuronal activation. For each training batch $B_b$,
we compute the mean activation of neuron $i$:
\begin{equation}
\bar{h}_{i,b}
=
\frac{1}{|B_b|}
\sum_{x\in B_b}h_i^{(S)}(x).
\end{equation}

The instantaneous Representation Flux is defined as
the absolute difference between consecutive batch means:
\begin{equation}
\Phi_{i,b}
=
\left|\bar{h}_{i,b}-\bar{h}_{i,b-1}\right|.
\end{equation}
Flux measures differences in mean neuronal activation between consecutive minibatches. Because these minibatches contain different examples, the signal reflects both changes in model representations and sampling variability, without isolating optimization-induced representational drift. We therefore use Flux as an empirical signal for structural adaptation rather than a direct measure of representational drift.

To suppress transient fluctuations, we maintain an
exponential moving average:
\begin{equation}
\widehat{\Phi}_{i,b}
=
\lambda_\Phi\widehat{\Phi}_{i,b-1}
+
(1-\lambda_\Phi)\Phi_{i,b},
\end{equation}
with $\lambda_\Phi=0.9$. Epoch-level Flux statistics
are used to determine persistent eligibility for
structural adaptation.

Low Flux alone does not imply redundancy, since
stable representations may remain functionally
important. We therefore additionally measure
neuronal activity:
\begin{equation}
A_{i,b}
=
\frac{1}{|B_b|}
\sum_{x\in B_b}|h_i^{(S)}(x)|,
\end{equation}
with exponential smoothing
\begin{equation}
\widehat{A}_{i,b}
=
\lambda_A\widehat{A}_{i,b-1}
+
(1-\lambda_A)A_{i,b},
\end{equation}
where $\lambda_A=0.9$. Neurons exhibiting
persistently low Flux and activity become
candidates for pruning.

\subsection{Flux-Driven Birth and Death}
\label{sec:birth_death}

\paragraph{Neuron birth.}
Structural adaptation is evaluated periodically after
a prescribed number of processed training examples. Flux and activity statistics are updated during
training, whereas structural changes are evaluated
periodically to reduce sensitivity to transient
fluctuations and avoid excessively frequent
architectural modifications.
At each structural check, neurons whose mean
Representation Flux over the preceding evaluation
interval exceeds $\theta_{\mathrm{birth}}$
accumulate a persistence counter, measured in
processed examples. Once the required persistence
threshold is reached, eligible neurons are ranked
by exponentially smoothed Flux, and those with
the highest values are selected for splitting.

Splitting creates a child neuron whose input and
incoming recurrent weights are copied from the
parent with small Gaussian perturbations. The
parent's outgoing recurrent weights and output-layer
contribution are divided equally between parent
and child. Two directed recurrent connections are
also introduced between them. Inspired by
function-preserving network expansion
\citep{chen2016net2net,wu2019splitting},
this operation aims to limit the immediate
functional perturbation while introducing
additional trainable capacity.

\paragraph{Neuron death.}
Pruning targets neurons exhibiting persistently low
activity and Flux. At each structural check,
candidates must fall within the lower population
quantiles of both statistics, with Flux additionally
constrained by an absolute upper threshold.
Qualifying neurons accumulate a persistence counter,
which resets whenever either criterion is violated.

Once the required persistence threshold is reached,
eligible neurons are ranked by exponentially
smoothed activity, and the least active are removed
first. Combining activity and Flux reduces the risk
of pruning strongly active neurons whose
representations have stabilized.

\paragraph{Population dynamics.}
Pruning precedes neurogenesis, subject to minimum
and maximum population constraints and per-check
event limits. The population evolves according to
\begin{equation}
N_{k+1}=N_k+B_k-D_k,
\end{equation}
where $k$ indexes structural checks and $B_k,D_k$
denote the numbers of neurons created and removed,
respectively. No target population size is prescribed.

\subsection{Developmental Control Parameters}
\label{sec:developmental_parameters}

We distinguish architectural initial conditions
$(N_0,\beta)$ from developmental controls governing
when and how rapidly structural adaptation occurs.

The birth threshold $\theta_{\mathrm{birth}}$
determines the minimum mean Representation Flux
required for a neuron to accumulate birth eligibility.
Pruning instead requires simultaneous membership in
the lower tails of the population's activity and Flux
distributions, defined by fractions $q_A$ and $q_\Phi$,
together with an absolute Flux bound
$\theta_{\mathrm{death}}$.
Birth and death persistence requirements,
$p_{\mathrm{birth}}$ and $p_{\mathrm{death}}$,
specify the cumulative number of processed examples
across consecutive qualifying structural checks.
Eligibility counters reset when their respective
conditions are no longer satisfied.

Structural decisions are evaluated after intervals
of $T_{\mathrm{check}}$ processed examples.
An initial warm-up period $T_{\mathrm{warm}}$
disables structural modifications while allowing
ordinary parameter optimization and accumulation
of neuronal statistics. Following a structural
change, a cooldown $T_{\mathrm{cool}}$ prevents
further modifications until the prescribed number
of examples has been processed.

The structural update rate $r$ specifies the nominal
number of births and deaths permitted at each structural
check. For fractional rates, the integer allowance $r_k$
alternates between the adjacent integers to achieve
the prescribed average rate. For example, $r=2.5$
corresponds to alternating allowances of two and three
events. The numbers of births and deaths at check $k$
therefore satisfy
\begin{equation}
B_k \leq r_k,\qquad D_k \leq r_k.
\end{equation}
The rate constrains the speed of architectural
development without determining which neurons
are selected.

Structural changes are additionally subject to
minimum and maximum population sizes,
$N_{\min}$ and $N_{\max}$, which prevent
unbounded growth or complete collapse without
prescribing a target architecture.
Most developmental parameters regulate the timing,
stability, or permissible extent of structural
adaptation rather than directly selecting neurons
for modification. The primary developmental controls
are the birth and pruning eligibility criteria and
the structural update rate $r$. Although temporal
safeguards and population bounds can influence
the resulting architecture, none explicitly prescribes
a target population size or architectural equilibrium.
Architectural development instead results from
the interaction between learned activation statistics
and local birth and pruning decisions.

\subsection{Datasets and Evaluation Protocols}
\label{sec:datasets_protocol}

Networks are trained across different initial sizes $N_0$
and random seeds. We track predictive performance,
Representation Flux, neuronal activity, and structural
dynamics, including neuron births and deaths. We investigate whether networks initialized with different
capacities converge toward a common late-training regime,
\begin{equation}
N(T;N_0^{(1)},\beta,D)
\approx
N(T;N_0^{(2)},\beta,D),
\end{equation}
where $\beta$ denotes initial connectivity and $D$ the task.

We use the term attractor-like architectural behavior
to describe reduced dependence of late-training neuron
counts on initial network size. In addition to the empirical study, our theoretical analysis
provides sufficient conditions for local population
stability (Appendix~\ref{app:architectural_equilibrium})
and initial-condition forgetting under contractive
population dynamics
(Appendix~\ref{app:initial_condition_forgetting}).

\paragraph{Reported quantities.}
We evaluate architectural development at three levels:
developmental dynamics, architectural properties,
and cross-run architectural similarity.

\textit{Developmental dynamics.}
We track training and test accuracy, active neuron count
$N(t)$, Representation Flux, and neuronal activity
throughout training. Initial, best-performing, and final
neuron counts are recorded to characterize architectural
development and population stabilization across
initializations and random seeds.

\textit{Architectural properties.}
Final architectures are characterized by neuron count,
recurrent connectivity density, and the spectral radius
of the masked, unnormalized recurrent weight matrix
\citep{jaeger2001echo,verstraeten2007experimental}.
Metrics are summarized across seeds using the mean and
standard deviation. 

\textit{Cross-run architectural similarity.}
We compare independently developed networks using
linear Centered Kernel Alignment (CKA) for
representational similarity \citep{kornblith2019similarity},
Hungarian-aligned neuron parameter-profile cosine
similarity \citep{kuhn1955hungarian}, and prediction
agreement for functional similarity.
Comparisons include runs with identical and different
initial network sizes.
Detailed metric definitions and statistical
procedures are provided in
Appendix~\ref{app:metrics}.

\paragraph{Datasets and evaluation.}
We investigate architectural development primarily on
Fashion-MNIST \citep{xiao2017fashion}, with MNIST
\citep{lecun1998gradient} as an additional benchmark.
For standalone image-classification experiments, images
are flattened and supplied directly to the neural
container without pretrained features or convolutional
layers. We evaluate generalization across image and text
modalities using CIFAR-10 \citep{krizhevsky2009learning}
and AG News \citep{zhang2015character}.
Finally, Split CIFAR-100 \citep{krizhevsky2009learning}
is used to assess adaptive capacity allocation in
task-incremental continual learning.

\paragraph{Training and model selection.}
Networks are trained using cross-entropy loss and Adam.
Structural updates are triggered after fixed intervals
of processed training examples. At each update, Flux
and activity statistics are aggregated, pruning is
performed before neurogenesis, and the Adam optimizer
is reinitialized whenever the network parameterization
changes. Classification accuracy is evaluated after
each training epoch.

We distinguish the best-performing checkpoint from
the final training epoch. Best-checkpoint accuracy is
reported descriptively because retrospective selection
using test accuracy introduces optimistic bias.
Final-epoch architectures are used to characterize
the outcome of the complete developmental trajectory.
Training continues beyond peak predictive performance
to examine late-training architectural stabilization.
Network size and cumulative birth and death events
are recorded alongside predictive performance.

\paragraph{Generality}
We investigate whether the developmental dynamics of
FlowGenesis depend on the initial number of hidden
neurons, $N_0$, and whether the observed architectural
self-organization generalizes across datasets and
modalities. We consider Fashion-MNIST with
$N_0\in\{20,40,\allowbreak 80,160\}$,
CIFAR-10 with
$N_0\in\{50,100,\allowbreak 200,400\}$,
and AG News with
$N_0\in\{20,40,\allowbreak 80,160\}$.
 
Each configuration is evaluated across five random
seeds, $\{0,1,2,3,4\}$.
These experiments span an eightfold range of initial
network sizes on each dataset. We track network size and test accuracy throughout training, quantify final structural properties, and compare the resulting networks using weight similarity, representational similarity, and prediction agreement.

\paragraph{Ablation protocol.}
We investigate the sensitivity of
architectural development to initial network size,
recurrent connectivity, structural rate, warm-up
duration, and birth and death criteria on Fashion-MNIST.
We additionally compare full bidirectional adaptation
against birth-only, death-only, and static architectures.
Generalization is examined on CIFAR-10 and AG News,
while continual learning is evaluated separately on
Split CIFAR-100. Detailed parameter grids and
experimental configurations are provided in
Appendix~\ref{app:ablation_protocol}.

\paragraph{Continual learning.}
We evaluate FlowGenesis in a progressive
task-incremental continual-learning setting using
Split CIFAR-100. Each task introduces a new set
of classes, with previously learned modules frozen
and a new module initialized for the incoming task.
FlowGenesis regulates neuron birth and pruning
within each new module during training, while
preserving parameters learned for earlier tasks.

We compare adaptive growth against fixed-capacity
progressive expansion, evaluating classification
accuracy, cumulative parameter count, and neuron
allocation across the task sequence.

\section{Results}

\subsection{Robustness to Initial Network Size}
\label{sec:n0_ablation}

\begin{table}[t]
\centering
\caption{
Architectural development and final topology of FlowGenesis
across Fashion-MNIST, CIFAR-10, and AG News for different
initial network sizes $N_0$.
$N_{\mathrm{best}}$ and $N_f$ denote neuron counts at the
best-performing checkpoint and at the end of training,
respectively. Births and deaths are cumulative structural
events, $E_f$ is the final number of recurrent connections,
and density is the fraction of active recurrent connections.
Values are mean $\pm$ standard deviation across five seeds.
}
\label{tab:n0_structure}

\begingroup
\scriptsize
\setlength{\tabcolsep}{3pt}
\renewcommand{\arraystretch}{1.15}

\resizebox{\linewidth}{!}{%
\begin{tabular}{@{}l c c c c c c c@{}}
\toprule
Dataset
& $N_0$
& $N_{\mathrm{best}}$
& $N_f$
& Births
& Deaths
& $E_f$
& Density \\
\midrule

\multicolumn{8}{c}{\textbf{Fashion-MNIST}} \\
\midrule
& 20
& $52.8\pm3.0$
& $61.6\pm10.1$
& $512.4\pm53.1$
& $470.8\pm48.4$
& $1449.4\pm424.3$
& $0.379\pm0.039$ \\

& 40
& $56.4\pm4.2$
& $53.0\pm5.0$
& $433.2\pm22.6$
& $420.2\pm20.4$
& $837.2\pm170.8$
& $0.298\pm0.042$ \\

& 80
& $50.6\pm3.5$
& $55.4\pm5.4$
& $365.0\pm43.6$
& $389.6\pm45.4$
& $887.0\pm175.7$
& $0.287\pm0.024$ \\

& 160
& $51.8\pm6.6$
& $51.4\pm4.2$
& $295.8\pm29.3$
& $404.4\pm31.4$
& $631.2\pm168.5$
& $0.235\pm0.031$ \\

\midrule
\multicolumn{8}{c}{\textbf{CIFAR-10}} \\
\midrule
& 50
& $96.0\pm38.5$
& $60.2\pm2.3$
& $1214.4\pm353.5$
& $1204.2\pm351.5$
& $1129.4\pm226.7$
& $0.311\pm0.059$ \\

& 100
& $103.4\pm30.3$
& $60.6\pm7.8$
& $1205.0\pm267.7$
& $1244.4\pm261.2$
& $887.8\pm232.4$
& $0.241\pm0.037$ \\

& 200
& $134.4\pm79.7$
& $62.6\pm4.8$
& $1170.2\pm399.0$
& $1307.6\pm395.0$
& $1073.6\pm171.0$
& $0.273\pm0.024$ \\

& 400
& $181.6\pm123.4$
& $64.0\pm6.8$
& $1195.0\pm296.0$
& $1531.0\pm292.1$
& $916.2\pm242.0$
& $0.222\pm0.032$ \\

\midrule
\multicolumn{8}{c}{\textbf{AG News}} \\
\midrule
& 20
& $77.2\pm25.1$
& $60.2\pm3.8$
& $451.2\pm65.0$
& $411.0\pm63.7$
& $943.8\pm227.0$
& $0.261\pm0.059$ \\

& 40
& $67.4\pm7.4$
& $63.8\pm7.4$
& $395.2\pm26.5$
& $371.4\pm25.8$
& $945.2\pm233.8$
& $0.234\pm0.057$ \\

& 80
& $59.4\pm5.3$
& $57.0\pm2.8$
& $211.4\pm48.0$
& $234.4\pm45.8$
& $695.8\pm98.6$
& $0.214\pm0.021$ \\

& 160
& $62.6\pm5.5$
& $58.4\pm2.7$
& $164.6\pm40.4$
& $266.2\pm39.0$
& $823.8\pm113.8$
& $0.241\pm0.017$ \\

\bottomrule
\end{tabular}%
}
\endgroup

\end{table}

\begin{table}[t]
\centering
\caption{
Predictive performance, recurrent dynamics, and
cross-initialization similarity of FlowGenesis across
Fashion-MNIST, CIFAR-10, and AG News.
Best-checkpoint and final-epoch test accuracies are
reported in percent.
The spectral radius $\rho$ is computed from the masked,
unnormalized recurrent weight matrix rather than the
degree-normalized operator used during network evaluation.
Weight similarity is computed after Hungarian alignment;
CKA measures representational similarity, and agreement
measures prediction consistency across different initial
network sizes.
Values are mean $\pm$ standard deviation across five seeds.
}
\label{tab:n0_similarity}

\begingroup
\scriptsize
\setlength{\tabcolsep}{3pt}
\renewcommand{\arraystretch}{1.15}

\resizebox{\linewidth}{!}{%
\begin{tabular}{@{}l c c c c c c c@{}}
\toprule
Dataset
& $N_0$
& Best acc. (\%)
& Final acc. (\%)
& $\rho$
& Cross-$N_0$ weight sim.
& Cross-$N_0$ CKA
& Agreement (\%) \\
\midrule

\multicolumn{8}{c}{\textbf{Fashion-MNIST}} \\
\midrule
& 20
& $87.60\pm0.38$
& $86.97\pm0.61$
& $0.526\pm0.101$
& $0.433\pm0.005$
& $0.884\pm0.015$
& $90.9\pm0.4$ \\

& 40
& $87.66\pm0.20$
& $86.69\pm0.68$
& $0.648\pm0.173$
& $0.440\pm0.011$
& $0.893\pm0.011$
& $90.8\pm0.9$ \\

& 80
& $87.60\pm0.29$
& $86.94\pm0.57$
& $0.592\pm0.151$
& $0.434\pm0.018$
& $0.889\pm0.005$
& $91.0\pm0.8$ \\

& 160
& $87.64\pm0.14$
& $86.66\pm0.29$
& $0.699\pm0.149$
& $0.440\pm0.007$
& $0.883\pm0.029$
& $90.9\pm0.5$ \\

\midrule
\multicolumn{8}{c}{\textbf{CIFAR-10}} \\
\midrule
& 50
& $51.59\pm0.32$
& $49.16\pm0.57$
& $0.807\pm0.483$
& $0.415\pm0.014$
& $0.628\pm0.030$
& $56.1\pm1.4$ \\

& 100
& $51.13\pm0.25$
& $48.13\pm1.49$
& $0.708\pm0.094$
& $0.419\pm0.019$
& $0.651\pm0.037$
& $55.0\pm3.6$ \\

& 200
& $51.30\pm0.26$
& $49.03\pm0.31$
& $0.659\pm0.069$
& $0.412\pm0.008$
& $0.630\pm0.039$
& $56.3\pm0.7$ \\

& 400
& $51.56\pm0.35$
& $49.29\pm0.66$
& $0.694\pm0.140$
& $0.418\pm0.007$
& $0.651\pm0.023$
& $56.5\pm1.0$ \\

\midrule
\multicolumn{8}{c}{\textbf{AG News}} \\
\midrule
& 20
& $89.77\pm0.09$
& $88.97\pm0.09$
& $5.881\pm1.219$
& $0.660\pm0.012$
& $0.827\pm0.007$
& $84.4\pm0.7$ \\

& 40
& $89.80\pm0.05$
& $89.19\pm0.16$
& $4.129\pm0.827$
& $0.666\pm0.011$
& $0.819\pm0.013$
& $84.4\pm0.3$ \\

& 80
& $89.82\pm0.10$
& $89.16\pm0.12$
& $3.790\pm0.316$
& $0.663\pm0.008$
& $0.817\pm0.015$
& $84.8\pm0.4$ \\

& 160
& $89.65\pm0.08$
& $89.20\pm0.18$
& $3.977\pm1.106$
& $0.661\pm0.013$
& $0.823\pm0.010$
& $84.7\pm0.6$ \\

\bottomrule
\end{tabular}%
}
\endgroup

\end{table}

We investigate the dependence of architectural
development on initial network size across
Fashion-MNIST, CIFAR-10, and AG News.
The experiments cover an eightfold range of initial
network sizes on each dataset.

\begin{figure*}[p]
    \centering
    \setlength{\tabcolsep}{2pt}

    \begin{minipage}[t]{0.49\textwidth}
        \centering
        \includegraphics[
            width=\linewidth,
            height=0.18\textheight,
            keepaspectratio
        ]{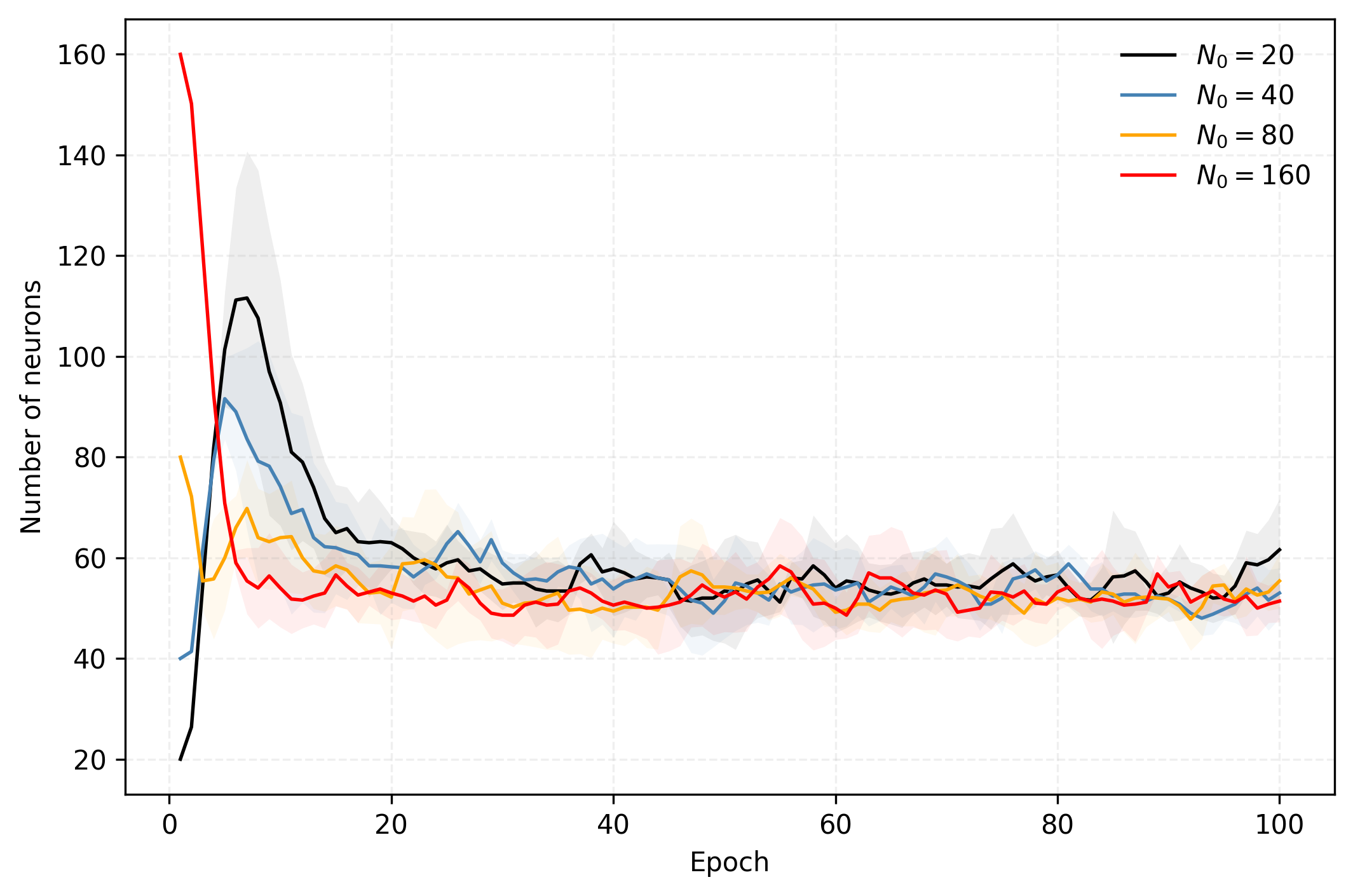}

        {\footnotesize (a) Fashion-MNIST: neurons}
    \end{minipage}\hfill
    \begin{minipage}[t]{0.49\textwidth}
        \centering
        \includegraphics[
            width=\linewidth,
            height=0.18\textheight,
            keepaspectratio
        ]{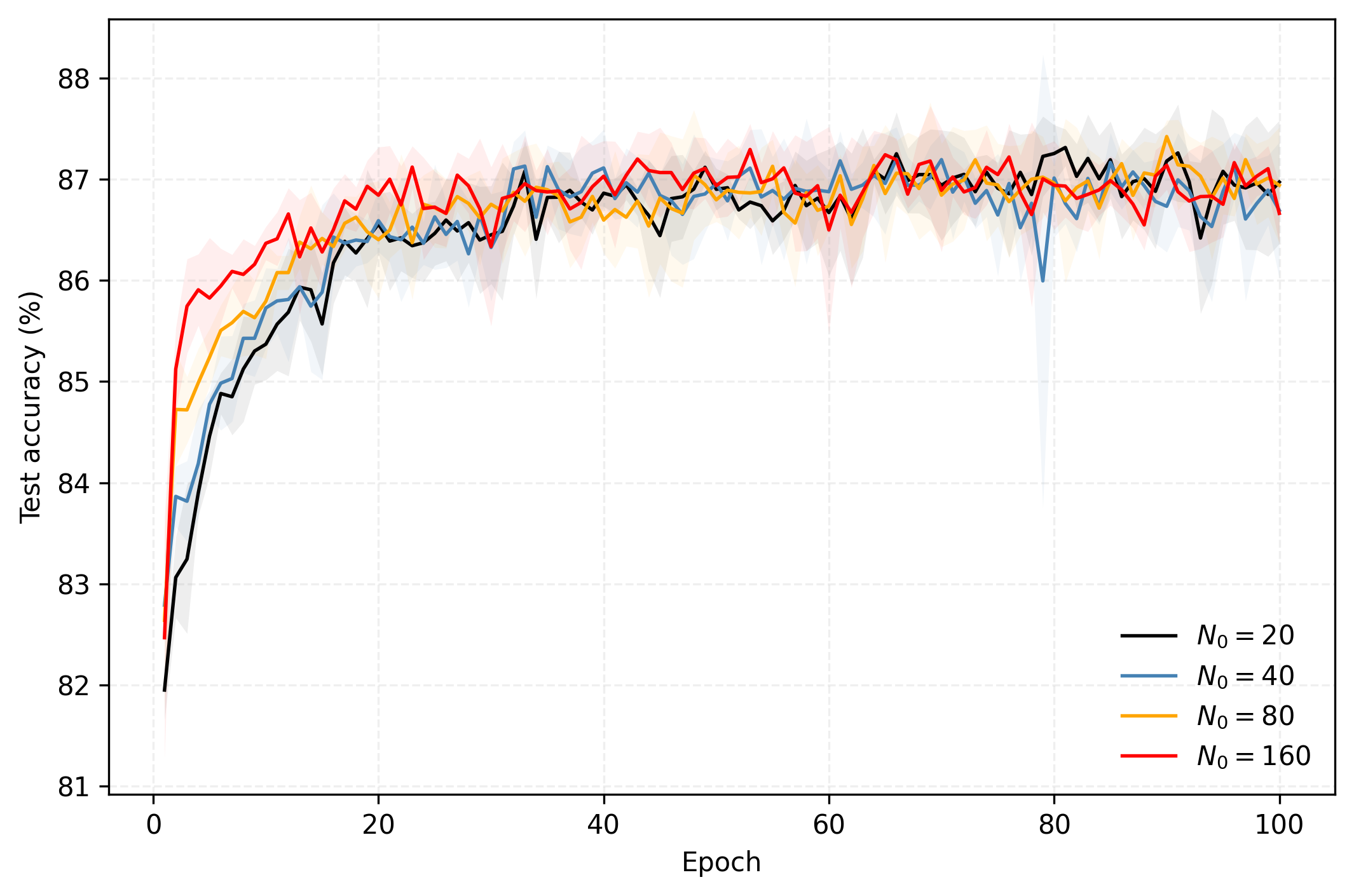}

        {\footnotesize (b) Fashion-MNIST: accuracy}
    \end{minipage}

    \vspace{0.2em}

    \begin{minipage}[t]{0.49\textwidth}
        \centering
        \includegraphics[
            width=\linewidth,
            height=0.18\textheight,
            keepaspectratio
        ]{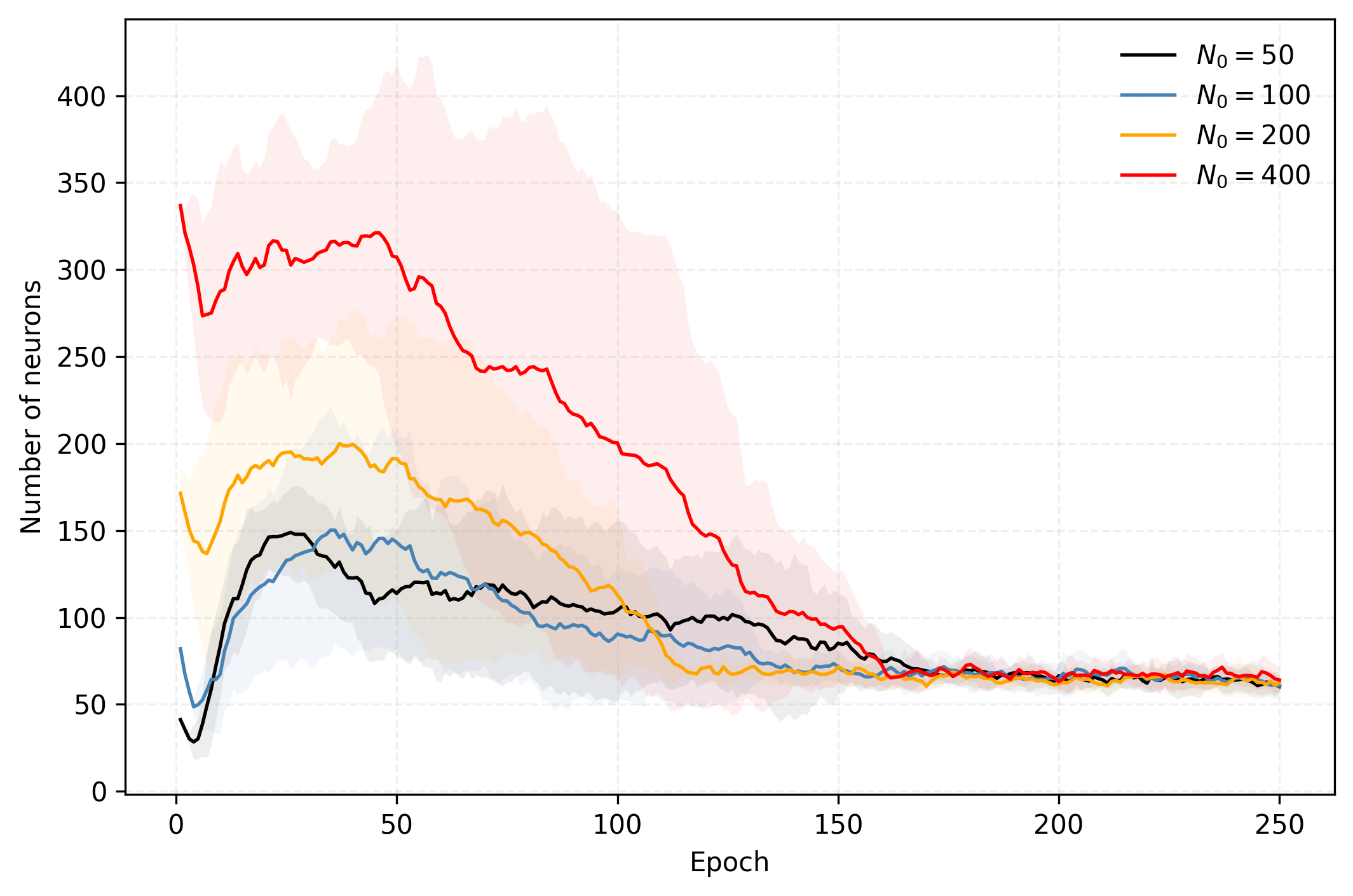}

        {\footnotesize (c) CIFAR-10: neurons}
    \end{minipage}\hfill
    \begin{minipage}[t]{0.49\textwidth}
        \centering
        \includegraphics[
            width=\linewidth,
            height=0.18\textheight,
            keepaspectratio
        ]{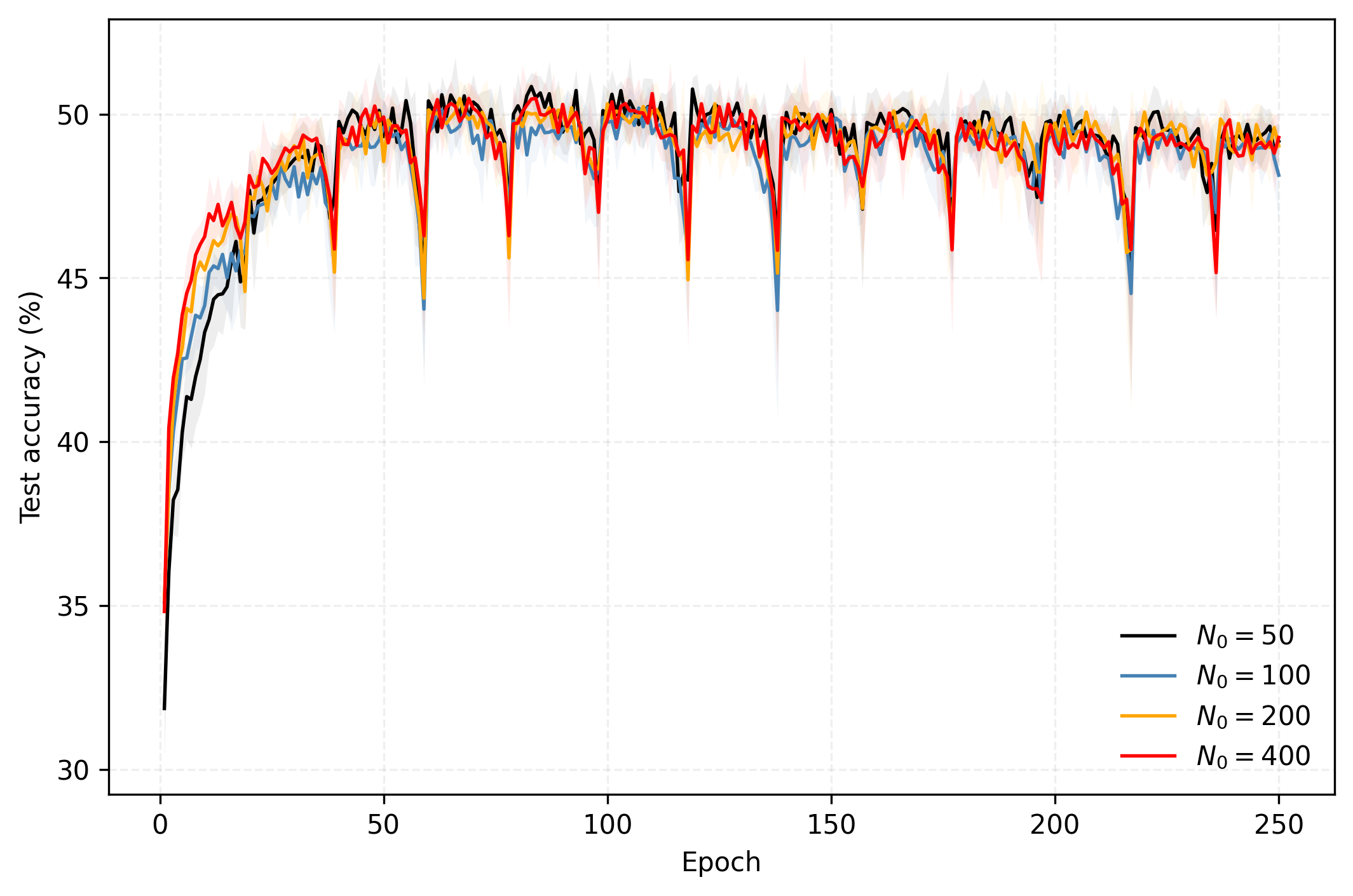}

        {\footnotesize (d) CIFAR-10: accuracy}
    \end{minipage}

    \vspace{0.2em}

    \begin{minipage}[t]{0.49\textwidth}
        \centering
        \includegraphics[
            width=\linewidth,
            height=0.18\textheight,
            keepaspectratio
        ]{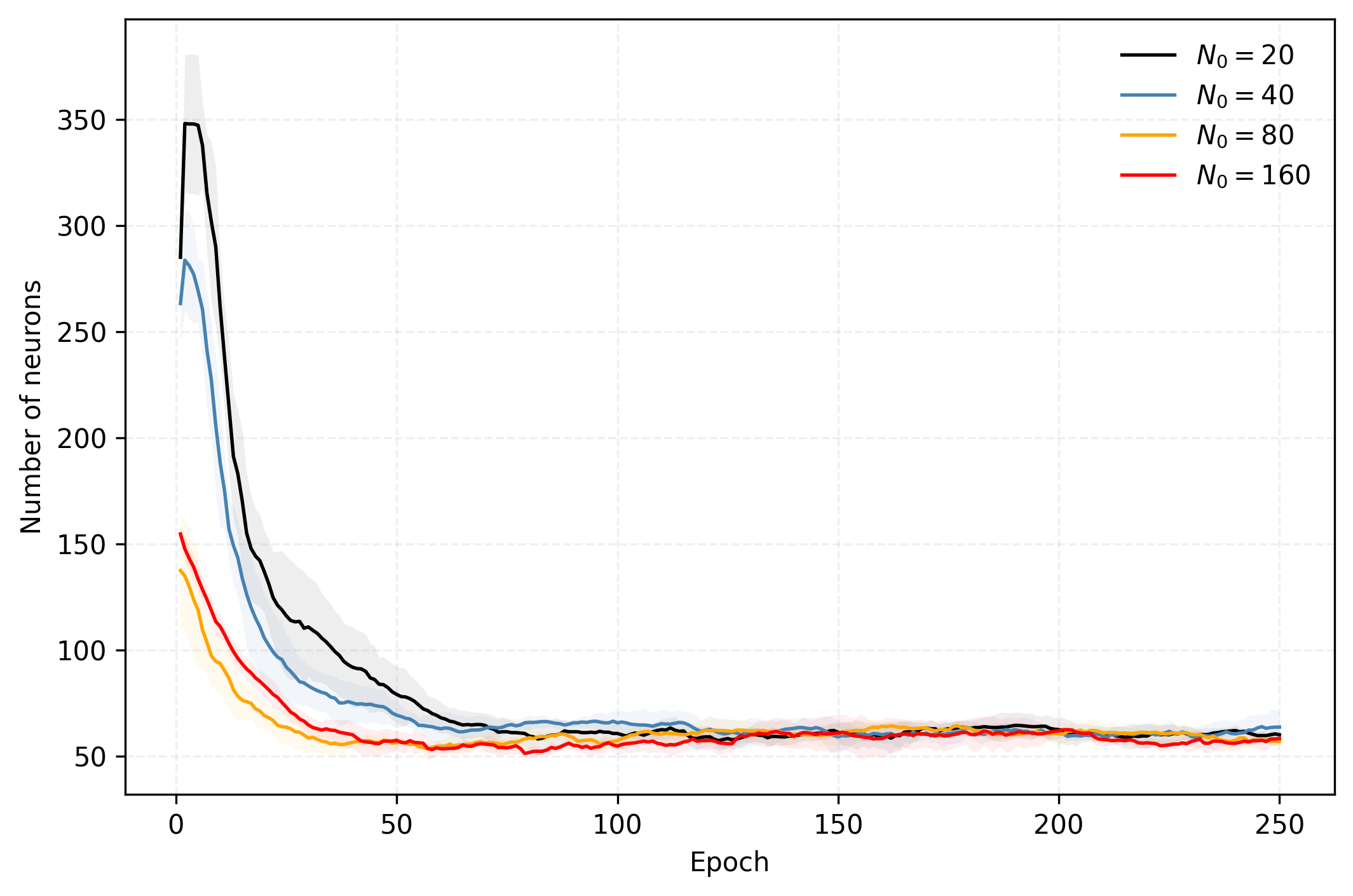}

        {\footnotesize (e) AG News: neurons}
    \end{minipage}\hfill
    \begin{minipage}[t]{0.49\textwidth}
        \centering
        \includegraphics[
            width=\linewidth,
            height=0.18\textheight,
            keepaspectratio
        ]{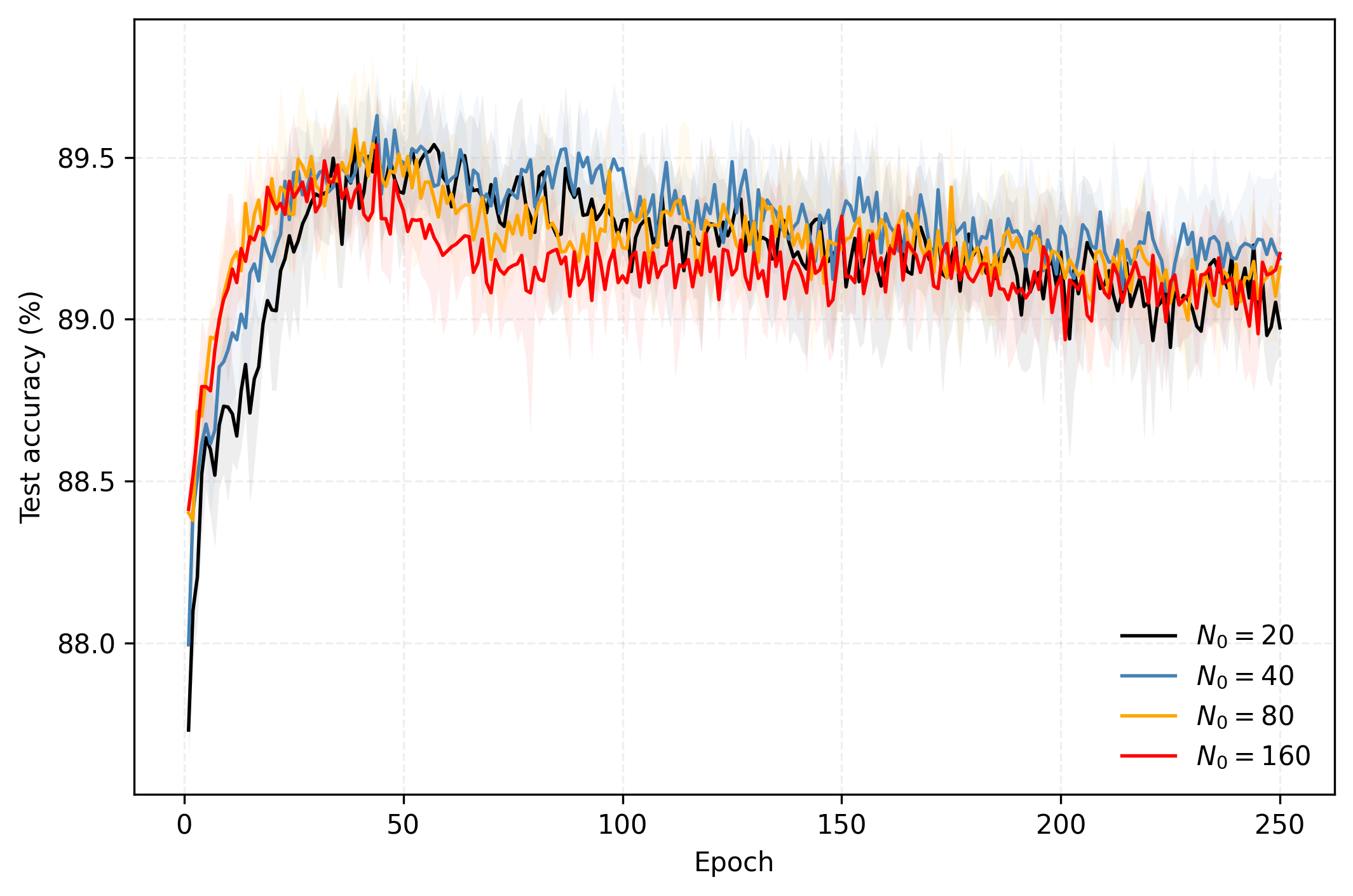}

        {\footnotesize (f) AG News: accuracy}
    \end{minipage}

    \vspace{-0.5em}

\caption{
Effect of initial network size $N_0$ on FlowGenesis
architectural development and predictive performance
across Fashion-MNIST, CIFAR-10, and AG News.
Curves show means across five random seeds;
shaded regions indicate $\pm$ one standard deviation.
Fashion-MNIST uses
$N_0\in\{20,40,80,160\}$,
CIFAR-10 uses
$N_0\in\{50,100,200,400\}$,
and AG News uses
$N_0\in\{20,40,80,160\}$.
Left panels (a,c,e) show hidden-neuron populations;
right panels (b,d,f) show test accuracy.
Across datasets, final neuron counts approach
similar ranges despite differences in initial
network size.
}

    \label{fig:n0_dynamics}
\end{figure*}
\paragraph{Convergence of network size across datasets.}
Figure~\ref{fig:n0_dynamics} shows that substantially different initial network sizes produce distinct developmental trajectories, but these trajectories approach relatively narrow ranges of final neuron counts.

On Fashion-MNIST, small networks undergo pronounced transient expansion, whereas larger initializations contract. Nevertheless, the final mean network sizes are $61.6\pm10.1$, $53.0\pm5.0$, $55.4\pm5.4$, and $51.4\pm4.2$ for $N_0=20,40,80,160$, respectively. Despite the eightfold variation in initialization, all four groups terminate with approximately 50–62 neurons. 

CIFAR-10 exhibits a similar phenomenon despite substantially different transient dynamics. The mean neuron counts at the best-performing checkpoints increase from $96.0\pm38.5$ at $N_0=50$ to $181.6\pm123.4$ at $N_0=400$, indicating considerable variability in the architectures encountered during training. However, the final populations reach mean sizes of
$60.2\pm2.3$, $60.6\pm7.8$, $62.6\pm4.8$,
and $64.0\pm6.8$ neurons, respectively.  Thus, substantially different developmental histories can lead to similar final network sizes.

On AG News, initial populations spanning an eightfold
range approach similar final mean neuron counts of
$60.2\pm3.8$, $63.8\pm7.4$, $57.0\pm2.8$,
and $58.4\pm2.7$ for $N_0=20,40,80,160$,
respectively. Thus, the final mean populations
remain between $57.0$ and $63.8$ neurons despite
substantial differences in initialization.

Taken together, these results reveal a recurring macroscopic pattern across all three datasets: although the initial populations differ substantially, the developmental process produces final networks occupying comparatively narrow ranges of neuron counts. In these experiments, the ranges overlap around approximately 50--65 neurons. This numerical similarity should not be interpreted as a universal preferred network size, since the equilibrium population may depend on task difficulty, model configuration, structural update rules, and training duration. The more general observation is the relative insensitivity of final population size to initialization within each experimental setting.

\paragraph{Predictive performance and initialization robustness.}
The emergence of similar final network sizes is
accompanied by relatively stable test accuracy
across different initializations. On Fashion-MNIST, mean best-checkpoint test accuracy
ranges from $87.60\%$ to $87.66\%$, while mean
final-epoch test accuracy ranges from $86.66\%$
to $86.97\%$. Networks with smaller initial
populations exhibit slower early learning in some
configurations but eventually achieve test accuracy
comparable to that of larger initializations. On CIFAR-10, mean best-checkpoint test accuracy
ranges from $51.13\%$ to $51.59\%$, while mean
final-epoch test accuracy ranges from $48.13\%$
to $49.29\%$. These relatively narrow ranges
are observed despite substantial differences
in developmental trajectories. A similar pattern emerges on AG News, where mean
best-checkpoint test accuracy ranges from $89.65\%$
to $89.82\%$ across four initializations spanning
an eightfold difference in initial network size.

\paragraph{Structural remodeling and residual differences.}
Convergence in neuron count does not imply convergence
to an identical network topology. Across all three
datasets, FlowGenesis exhibits extensive structural
remodeling through neuron births and deaths, with
developmental trajectories depending on initialization. On Fashion-MNIST, mean final recurrent connectivity
density decreases from $0.379\pm0.039$ at $N_0=20$
to $0.235\pm0.031$ at $N_0=160$, indicating that
similar final neuron counts can coexist with
substantially different connectivity densities. On CIFAR-10, mean final recurrent edge counts range
from approximately $888$ to $1129$, while mean
connectivity densities range from $0.222$ to $0.311$.
Thus, networks approaching similar final neuron
counts retain differences in recurrent connectivity. AG News exhibits substantial remodeling despite
relatively stable final network sizes. As $N_0$
increases from $20$ to $160$, cumulative neuron
births decrease from $451.2\pm65.0$ to
$164.6\pm40.4$, while cumulative deaths decrease
from $411.0\pm63.7$ to $266.2\pm39.0$.
Across these initializations, mean final recurrent
edge counts range from approximately $696$ to $945$,
and connectivity densities range from $0.214$
to $0.261$. Together, these results distinguish convergence
in macroscopic network size from convergence
in detailed connectivity. Initial conditions
influence structural remodeling and aspects of
the final topology even when final neuron counts
and predictive performance are similar.
The observed behavior is therefore consistent
with the emergence of a family of related
architectural states rather than a unique final graph.

\paragraph{Within- and cross-initialization functional similarity.}
To assess whether independently developed architectures
share functional properties, we compare models trained
with different random seeds at the same initial network
size (\emph{within-$N_0$}) and across different initial
sizes (\emph{cross-$N_0$}). We measure Hungarian-aligned
recurrent weight similarity, representational similarity
using centered kernel alignment (CKA), and prediction
agreement. Pairwise similarities are aggregated at the
model level before computing group statistics. On Fashion-MNIST, within-$N_0$ CKA ranges from
$0.870$ to $0.906$, compared with $0.883$--$0.893$
across initializations. Prediction agreement is
$90.3$--$91.5\%$ within initializations and
$90.8$--$91.0\%$ across them, while weight similarity
ranges from $0.410$--$0.453$ and $0.433$--$0.440$,
respectively. These overlapping ranges suggest that
variation in initial network size does not introduce
markedly greater representational or predictive
differences than variation in random seed alone. On CIFAR-10, cross-initialization CKA ranges from
$0.628$ to $0.651$, prediction agreement from
$55.0\%$ to $56.5\%$, and weight similarity from
$0.412$ to $0.419$. Although absolute similarity
is lower than on Fashion-MNIST, the metrics vary
relatively little across initializations. The lower
prediction agreement should also be interpreted
in light of the lower classification accuracy. On AG News, cross-initialization CKA ranges from
$0.817$ to $0.827$, with prediction agreement
between $84.4\%$ and $84.8\%$, extending the
observed pattern to text classification. Overall, these results indicate that similar final
network sizes are accompanied by relatively consistent
representational similarity and prediction agreement
across initializations. This supports descriptive
functional robustness without implying identical
network topologies, representations, or decision
boundaries.
\paragraph{Generalization of architectural self-organization.}
Across Fashion-MNIST, CIFAR-10, and AG News, FlowGenesis
exhibits convergence toward relatively narrow final
neuron-count ranges despite substantial differences in
initial network size and developmental trajectories.
This convergence is primarily macroscopic: final
architectures retain differences in connectivity.
Nevertheless, predictive performance and
cross-initialization representational similarity
remain comparatively stable. These findings suggest that FlowGenesis develops
families of related architectural states rather than
converging to a unique network structure.
The observed behavior suggests attractor-like
self-organization, as networks with different initial
sizes approach similar final neuron counts.
However, these results do not prove the existence
of a dynamical attractor or establish that different
initializations produce statistically equivalent outcomes. Additional visualizations reveal dataset-dependent
patterns of neuronal response correlations, with
more pronounced functional clusters on Fashion-MNIST
and AG News and weaker clustering on CIFAR-10
(Appendix~\ref{app:architecture_visualization}).

\subsection{Neuron Birth and Death Ablations}
\label{sec:regime_ablation}

We examine the effects of disabling neuron birth
or death on Fashion-MNIST, using initial populations
$N_0\in\{20,40,80,160\}$ and five random seeds.
All other settings follow
Appendix~\ref{app:ablation_protocol}.

\begin{figure*}[t]
\centering

\begin{subfigure}[t]{0.49\textwidth}
\centering
\includegraphics[width=\linewidth]
{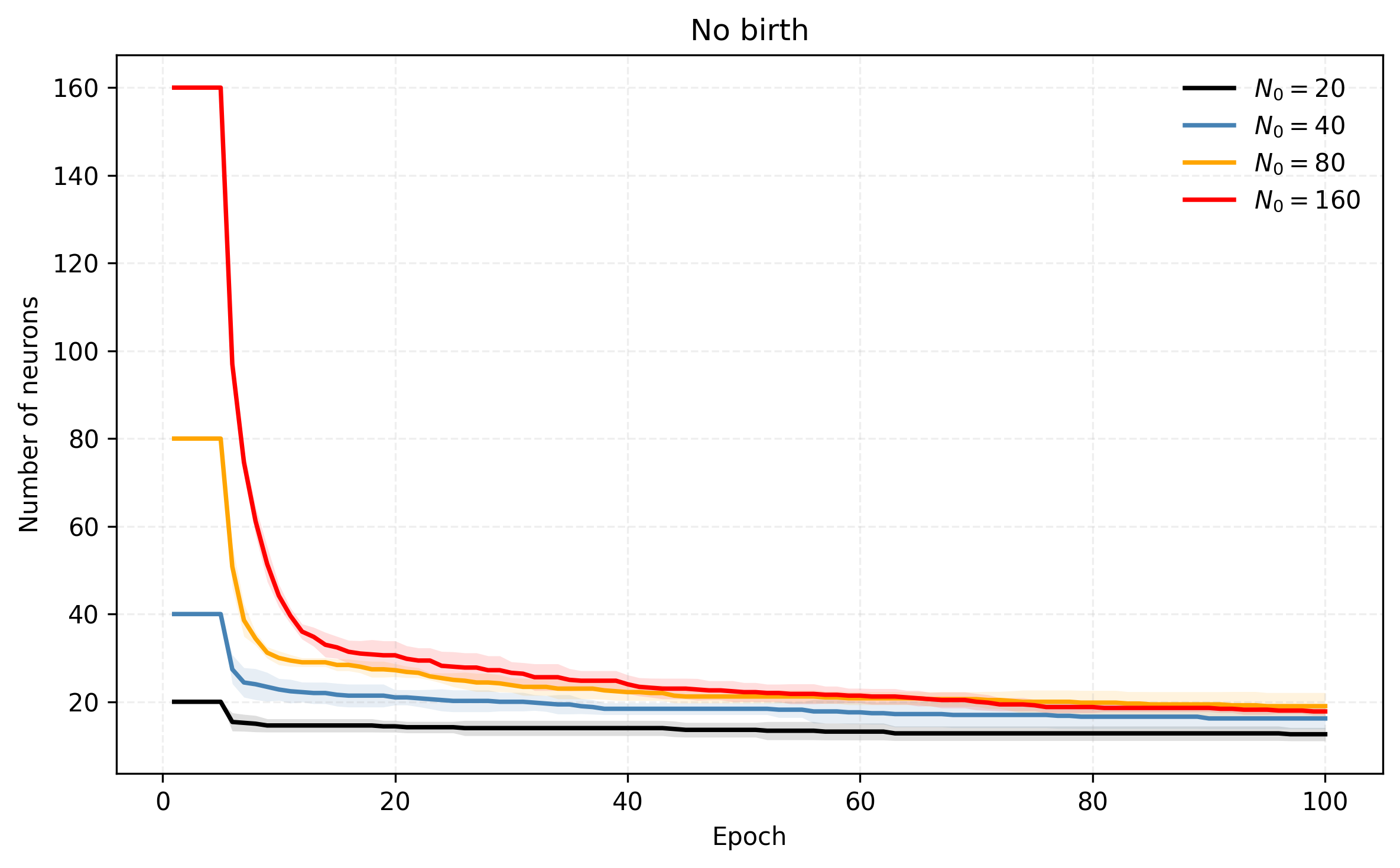}
\caption{Birth disabled (pruning only).}
\end{subfigure}
\hfill
\begin{subfigure}[t]{0.49\textwidth}
\centering
\includegraphics[width=\linewidth]
{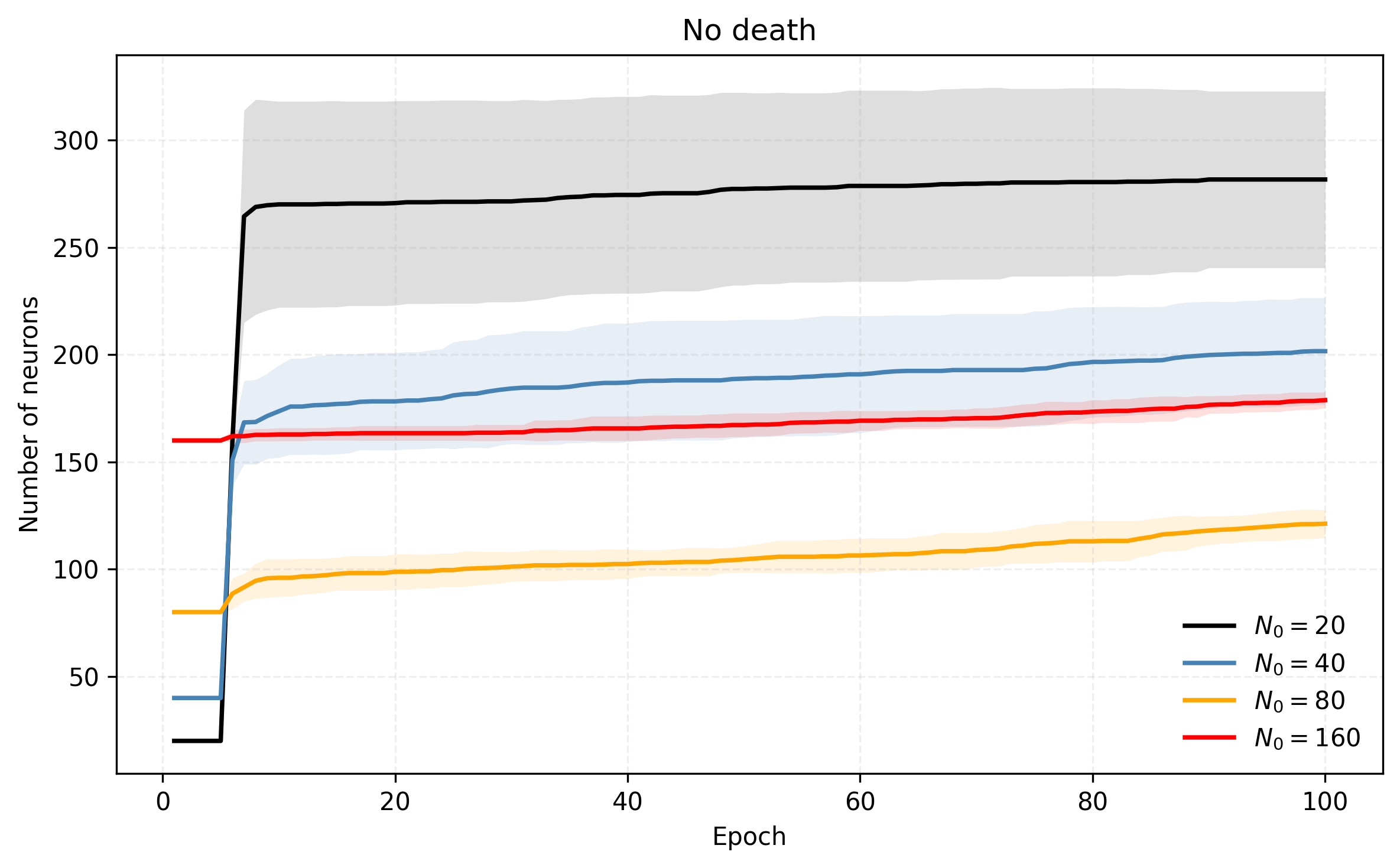}
\caption{Pruning disabled (birth only).}
\end{subfigure}

\caption{
Neuron population dynamics under structural ablations
on Fashion-MNIST for $N_0\in\{20,40,80,160\}$.
Disabling birth produces progressive contraction,
whereas disabling pruning permits population growth
without convergence to a common population size.
Curves show mean $\pm$ one standard deviation
across five random seeds.
}
\label{fig:regime_ablation}
\end{figure*}

Figure~\ref{fig:regime_ablation} shows that disabling
birth produces rapid population contraction, with
final sizes of approximately 13--19 neurons.
In contrast, disabling death leads to population
growth, with final sizes of approximately
121--282 neurons, depending on initialization.
Unlike the birth-disabled regime, these populations
remain substantially separated across initializations.

These results demonstrate complementary structural
roles: birth counteracts excessive contraction,
whereas death limits population expansion.
Neither mechanism alone reproduces the population
regulation observed when both are active.

\subsection{Sensitivity to Neuron Birth and Pruning Selection}
\label{sec:causal_selection}

We investigate the sensitivity of FlowGenesis to neuron
birth selection and pruning signals on Fashion-MNIST
($N_0=20$). The experiments examine whether the
neuron-specific information provided by Representation
Flux and activity contributes to architectural
adaptation. Birth and pruning mechanisms are varied
separately, while the remaining components of
FlowGenesis are kept unchanged.

\paragraph{Neuron birth selection.}
We compare three strategies:
\textbf{high-flux selection} (original FlowGenesis),
which prioritizes eligible neurons with the highest
exponentially smoothed Representation Flux;
\textbf{reverse-flux selection}, which prioritizes
the lowest-Flux neurons among eligible candidates;
and \textbf{unrestricted random selection},
which selects neurons randomly without enforcing
Flux-based birth eligibility.

\paragraph{Neuron pruning selection.}
We compare four strategies:
\textbf{original FlowGenesis}, which prioritizes
eligible neurons with persistently low activity
and Flux using exponentially smoothed activity;
\textbf{unrestricted random pruning}, which removes
activity- and Flux-based eligibility constraints;
\textbf{shuffled activity}, which permutes activity
values across neurons while preserving Flux;
and \textbf{shuffled Flux}, which permutes Flux
values while preserving activity.
In both shuffled conditions, pruning eligibility
and selection use the modified signals.

All strategies are evaluated across five paired seeds
with otherwise identical settings. Birth-selection
experiments retain the original pruning mechanism,
whereas pruning-selection experiments retain the
original birth mechanism.

\begin{table*}[t]
\centering
\caption{
Sensitivity to neuron birth and pruning selection on
Fashion-MNIST ($N_0=20$).
Mean $\pm$ standard deviation across five seeds.
Birth-selection experiments retain the original pruning
mechanism, while pruning-selection experiments retain
the original birth mechanism.
}
\label{tab:causal_selection}
\scriptsize
\setlength{\tabcolsep}{3pt}
\renewcommand{\arraystretch}{1.15}
\resizebox{\textwidth}{!}{%
\begin{tabular}{@{}lccccccc@{}}
\toprule
& \multicolumn{3}{c}{Birth selection}
& \multicolumn{4}{c}{Pruning selection} \\
\cmidrule(lr){2-4}
\cmidrule(l){5-8}
Metric
& High flux
& Reverse flux
& Random
& Original
& Random
& Shuffled activity
& Shuffled Flux \\
\midrule
Best acc. (\%)
& $87.61\pm0.37$
& $87.39\pm0.23$
& $88.40\pm0.16$
& $87.60\pm0.38$
& $86.39\pm0.19$
& $87.49\pm0.11$
& $87.51\pm0.10$ \\
Final acc. (\%)
& $86.99\pm0.59$
& $86.60\pm0.17$
& $87.86\pm0.41$
& $86.97\pm0.61$
& $85.79\pm0.36$
& $86.92\pm0.25$
& $86.68\pm0.54$ \\
Final neurons
& $57.2\pm11.3$
& $52.8\pm7.6$
& $140.8\pm8.5$
& $61.60\pm10.11$
& $85.20\pm7.85$
& $51.20\pm12.36$
& $55.00\pm4.36$ \\
\bottomrule
\end{tabular}%
}
\end{table*}

Table~\ref{tab:causal_selection} shows that
high-flux and reverse-flux selection yield
similar final network sizes
($57.2\pm11.3$ and $52.8\pm7.6$ neurons).
Unrestricted random selection achieves the
highest final accuracy ($87.86\pm0.41\%$)
but produces substantially larger networks
($140.8\pm8.5$ neurons).

Unrestricted random pruning increases final
network size from $61.6\pm10.1$ to
$85.2\pm7.9$ neurons while reducing final
accuracy from $86.97\pm0.61\%$ to
$85.79\pm0.36\%$.
Shuffling activity or Flux yields smaller
networks with comparable accuracy to the
original mechanism.

\subsection{Progressive Structural Adaptation in Continual Learning}
\label{sec:progressive_cl}

\paragraph{Experimental setting.}
We evaluate capacity efficiency in progressive continual learning on Split CIFAR-100, using 20 tasks of five classes or 50 tasks of two classes, with five random seeds. We compare Progressive FlowGenesis against a fixed-width baseline inspired by Progressive Neural Networks (PNN; \citealp{rusu2016progressive}). Both methods use a three-layer convolutional encoder (32, 64, 128 channels) producing 128-dimensional representations, trained on the first task, selected by validation accuracy, and subsequently frozen. Each task receives a recurrent module with sparse connectivity (initial connection probability 0.2), ten relaxation steps, a task-specific classifier, and 32-dimensional lateral representations from each preceding frozen module. FlowGenesis adapts module width through Flux-guided neuron birth and pruning, whereas PNN maintains a fixed width chosen to approximately match the early module sizes learned by FlowGenesis. Both methods use 40 training epochs per task, Adam optimization with learning rate $10^{-3}$, and batch size 128. For each task, the checkpoint with the highest validation accuracy is selected and subsequently frozen. FlowGenesis uses fixed structural adaptation criteria throughout the task sequence, without a prescribed target module width. Evaluation is task-incremental, with task identity available at inference.

\paragraph{Twenty-task results.}
As shown in Table~\ref{tab:progressive_cifar100},
FlowGenesis achieves $59.32\pm2.64\%$ average accuracy,
compared with $58.66\pm2.88\%$ for PNN
(uncorrected paired-test $p=0.0087$).
Simultaneously, cumulative active parameters decrease
from $0.650$M to $0.576$M, an $11.3\%$ reduction,
while cumulative neuron allocation decreases by $8.4\%$.
Thus, in the evaluated 20-task setting, FlowGenesis achieves higher observed accuracy and uses fewer active parameters than the fixed-width PNN baseline.

\begin{figure*}[t]
    \centering

    \includegraphics[
        width=\textwidth
    ]{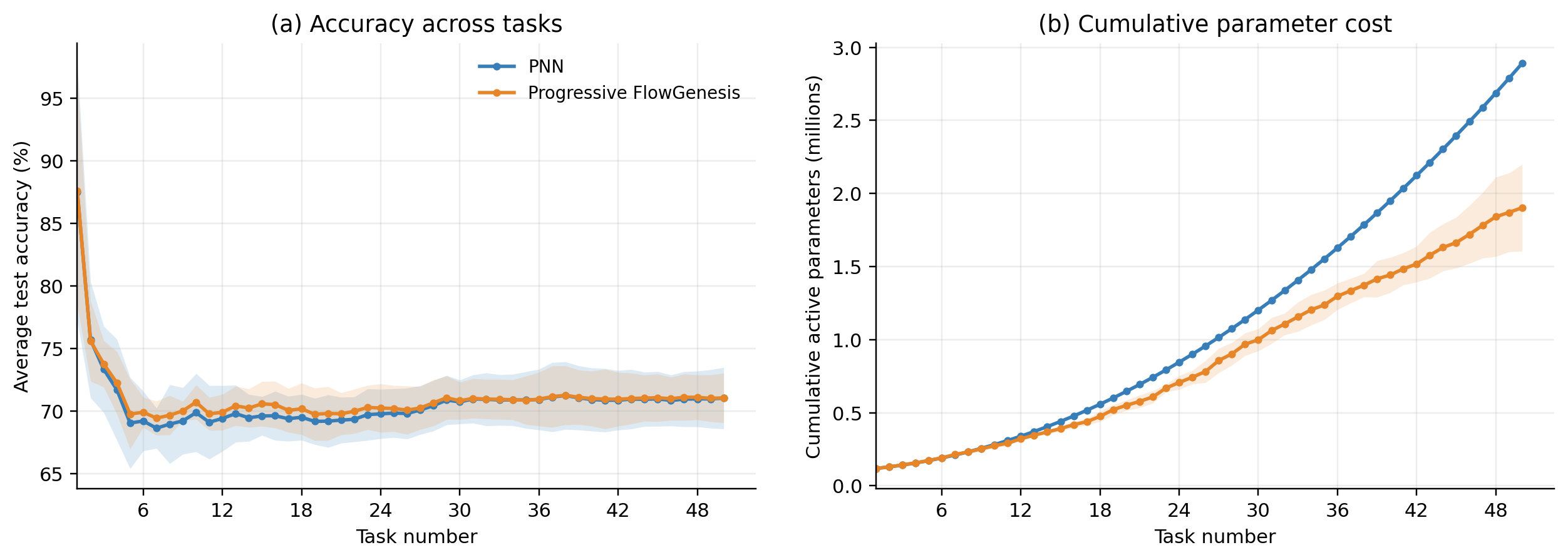}

    \vspace{0.5em}

    \includegraphics[
        width=\textwidth
    ]{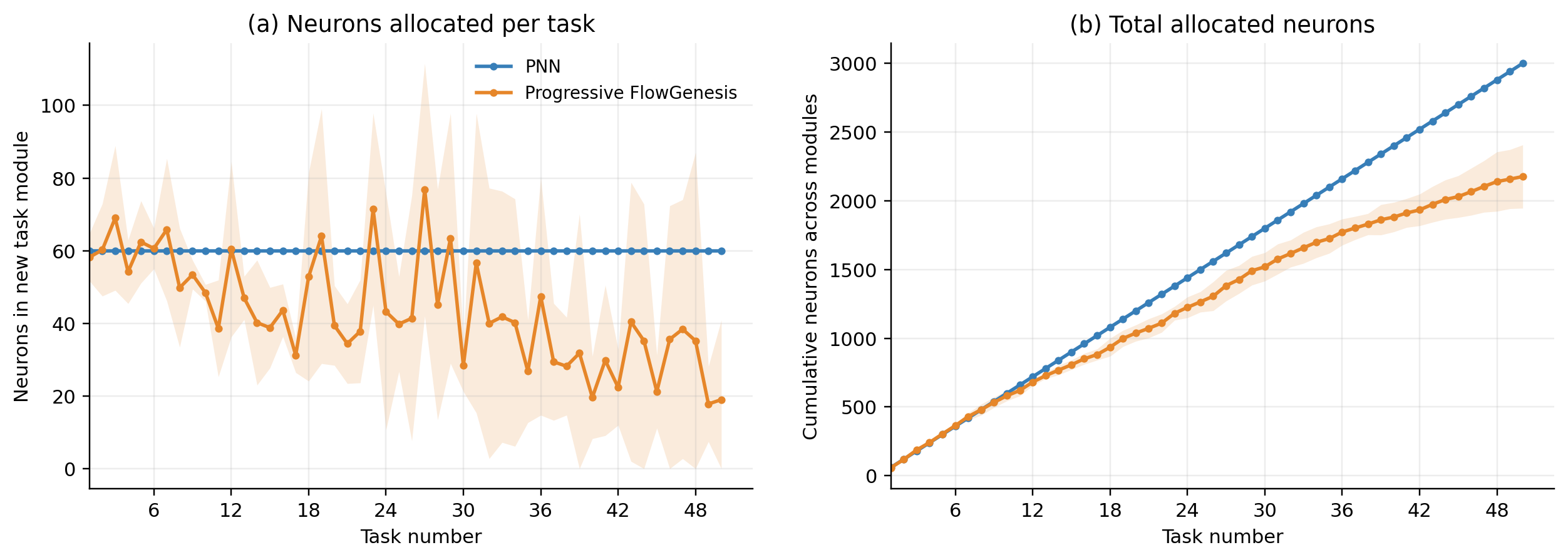}

    \caption{
    \textbf{Progressive FlowGenesis versus fixed-width
    Progressive Neural Networks (PNN) on Split CIFAR-100
    with 50 tasks (two classes per task).}
    \textbf{Top:} Average task-incremental test accuracy
    across all tasks observed so far (left) and cumulative
    active parameter count (right).
    \textbf{Bottom:} Number of neurons allocated to each
    new task module (left) and cumulative number of
    allocated neurons (right).
    Curves show means across five random seeds, with
    shaded regions indicating $\pm$ one sample standard
    deviation.
    PNN allocates 60 neurons per task, whereas FlowGenesis
    dynamically determines module size through
    representation-driven structural adaptation.
    At task 50, FlowGenesis achieves
    $71.05\pm1.98\%$ accuracy versus
    $71.03\pm2.46\%$ for PNN, while reducing
    cumulative active parameters from
    $2.89$M to $1.90$M (approximately 34\%).
    }
    \label{fig:progressive_flowgenesis}
\end{figure*}

\paragraph{Fifty-task results.}
For 50 tasks, FlowGenesis achieves
$71.05\pm1.98\%$ average accuracy versus
$71.03\pm2.46\%$ for PNN
(uncorrected paired-test $p=0.9718$).
Despite comparable accuracy, FlowGenesis reduces
cumulative active parameters from $2.890$M to
$1.903$M, a $34.2\%$ reduction
(uncorrected paired-test $p=0.0018$),
and cumulative neuron allocation from 3,000 to
2,177, a $27.4\%$ reduction
(uncorrected paired-test $p=0.0013$).
Figure~\ref{fig:progressive_flowgenesis} shows that
FlowGenesis allocates progressively smaller modules,
whereas PNN maintains a fixed width of 60 neurons.

\paragraph{Capacity allocation across task sequences.}
In the final task module, FlowGenesis allocates
$32.0\%$ fewer neurons than the fixed-width baseline
in the 20-task setting and $68.3\%$ fewer in the
50-task setting. Cumulative parameter savings reach
$11.3\%$ and $34.2\%$, respectively.
Although these results suggest increasing capacity
efficiency, differences in classes per task prevent
attributing the trend solely to sequence length.

\begin{table*}[t]
\centering
\caption{
Progressive FlowGenesis versus fixed-width Progressive Neural
Networks (PNN) on Split CIFAR-100.
Results are reported as mean $\pm$ standard deviation over
five paired seeds. Accuracy is evaluated in the
task-incremental setting and averaged over all tasks seen
at the final checkpoint. Active parameters include the
shared encoder and all allocated task modules.
$p$-values are from two-sided paired $t$-tests
(uncorrected).
}
\label{tab:progressive_cifar100}
\scriptsize
\setlength{\tabcolsep}{5pt}
\begin{tabular}{llccc}
\toprule
Setting & Metric
& PNN
& FlowGenesis
& $p$-value \\
\midrule
\multirow{3}{*}{20 tasks $\times$ 5 classes}
& Accuracy (\%)
& $58.66 \pm 2.88$
& $\mathbf{59.32 \pm 2.64}$
& $0.0087$ \\
& Active parameters (M)
& $0.650 \pm 0.0002$
& $\mathbf{0.576 \pm 0.022}$
& $0.0016$ \\
& Cumulative neurons
& $1200 \pm 0$
& $\mathbf{1099 \pm 37}$
& $0.0035$ \\
\midrule
\multirow{3}{*}{50 tasks $\times$ 2 classes}
& Accuracy (\%)
& $71.03 \pm 2.46$
& $71.05 \pm 1.98$
& $0.9718$ \\
& Active parameters (M)
& $2.890 \pm 0.0001$
& $\mathbf{1.903 \pm 0.297}$
& $0.0018$ \\
& Cumulative neurons
& $3000 \pm 0$
& $\mathbf{2177 \pm 230}$
& $0.0013$ \\
\bottomrule
\end{tabular}
\end{table*}

\section{Discussion}
\label{sec:discussion}

Our results demonstrate that FlowGenesis adapts network
architecture during training through neuron birth and
pruning guided by Representation Flux. Experiments on
image and text classification reveal attractor-like
architectural behavior: networks initialized with
substantially different capacities approach similar
final neuron-count ranges within each dataset.
Despite differences in connectivity and developmental
trajectories, these networks exhibit substantial
representational similarity and comparable predictive
performance. Developmental ablations further characterize this
behavior. The structural update rate and birth
eligibility threshold influence population dynamics,
whereas initial network size has relatively little
influence on final neuron count under the evaluated
conditions. Importantly, these attractor-like regimes
exhibit stable macroscopic population statistics
despite ongoing structural remodeling.
Neurons continue to be created and removed even after
population size has stabilized. 
The neuron-selection ablations reveal a trade-off
between predictive performance and network size.
Unrestricted random birth selection achieves higher
final accuracy than high-Flux selection but produces
substantially larger networks. Reverse-Flux selection
provides an additional comparison of birth priorities.
For pruning, shuffling activity or Flux yields
broadly comparable outcomes to the original
selection mechanism. These results indicate that
the observed architectural regulation does not
uniquely depend on neuron-specific Flux ranking.

\paragraph{Application to continual learning.}
Progressive FlowGenesis extends representation-driven
structural adaptation to sequential learning, enabling
adaptive capacity allocation based on learned
representations.
On Split CIFAR-100, FlowGenesis achieves accuracy
comparable to fixed-width Progressive Neural Networks
while allocating fewer neurons and parameters.
In the 50-task setting, cumulative parameter count
is reduced by 34.2\% and neuron count by 27.4\%,
with comparable predictive accuracy. These savings
are larger than those observed in the 20-task setting,
suggesting that adaptive capacity allocation becomes
increasingly beneficial over longer task sequences. More broadly, FlowGenesis provides a mechanism for
adaptive capacity allocation in continual learning
by preserving previously learned representations
and using Representation Flux to guide the allocation
of new neurons as tasks accumulate. In the current implementation, previously learned
modules remain frozen, preventing subsequent tasks
from overwriting their parameters. FlowGenesis
therefore addresses efficient capacity allocation
rather than catastrophic forgetting directly.
An important next step is to investigate whether
representation-driven structural adaptation can
support continual learning in shared networks
without explicit task boundaries.

\paragraph{Theoretical interpretation.}
Our theoretical analysis examines simplified
mechanisms of architectural self-organization.
Neuron splitting is designed to limit immediate
functional perturbations while introducing
additional trainable capacity. Exact functional
preservation is not guaranteed. Representation
Flux provides a local criterion for selecting
neurons for expansion based on differences in
activation statistics across consecutive minibatches. At the population level, a stationary mean-field
model admits equilibria when expected neuron birth
and pruning rates are equal. We derive sufficient
conditions for local asymptotic stability of these
equilibria. Under an additional contraction assumption, differences between population trajectories with
distinct initial neuron counts decrease geometrically,
up to a disturbance-dependent bound. These results characterize conditional properties
of reduced population dynamics. The corresponding
assumptions have not been established for the full
stochastic FlowGenesis algorithm. The theoretical
analyses are provided in
Appendix~\ref{app:flux_neurogenesis}.

\paragraph{Biological interpretation.}
FlowGenesis suggests a conceptual connection between
architectural self-organization and structural
plasticity in biological neural circuits.
Synaptic connections undergo formation and elimination
in response to experience
\citep{holtmaat2009experience}, while homeostatic
mechanisms help maintain stable neural activity
despite ongoing plasticity
\citep{turrigiano2012homeostatic}. These observations motivate investigating whether
biological neural circuits can maintain stable
macroscopic organization through continuous
structural remodeling rather than fixed connectivity.
We speculate that such organization could emerge
from local plasticity mechanisms without explicit
global control of network structure. Although
FlowGenesis does not reproduce the biological
processes underlying structural plasticity, it
provides a computational framework for studying
how local adaptive dynamics may produce stable
macroscopic organization.

\paragraph{Limitations and future directions.}
Representation Flux, estimated from consecutive
minibatch activation means, captures both
representational changes and sampling variability.
Consequently, it does not isolate
optimization-induced representational drift.
Developing more robust estimators and investigating
alternative local signals could improve structural
selection, potentially at the cost of greater
computational complexity. The recurrent neural container was deliberately
chosen for its structural flexibility, allowing
networks to grow and shrink through local neuron
birth and pruning. However, this flexibility comes
at the cost of architectural specialization and
potentially computational efficiency. Extending
FlowGenesis to convolutional and Transformer
architectures would help establish the generality
and scalability of the proposed mechanism.
Furthermore, the continual-learning experiments
are restricted to the task-incremental setting
with known task boundaries. Finally, attractor-like architectural behavior is
observed empirically over finite training horizons.
Although our theoretical analysis establishes
conditional results for population stability and
initial-condition forgetting, the existence,
stability, and invariance of attractors under
the complete stochastic, nonconvex training
dynamics remain open questions. Further analysis
could establish when local structural adaptation
produces stable macroscopic regimes and determine
how these regimes depend on task complexity,
training duration, and developmental parameters.

\section{Conclusion}
\label{sec:conclusion}

We introduced FlowGenesis, a framework in which neural architectures develop through representation-driven neuron birth and pruning. Experiments on image and text classification show that networks initialized with different sizes converge toward similar dataset-specific architectural regimes while maintaining comparable predictive performance. These attractor-like regimes exhibit similar macroscopic properties across initializations despite ongoing structural remodeling. The same principle enables adaptive capacity allocation in continual learning, reducing architectural expansion while achieving comparable observed accuracy. Together, these findings suggest that network architecture need not be fully specified before training, but can emerge and adapt through the learning process itself.

\newpage
\appendix

\section{Architecture Metrics and Statistical Analysis}
\label{app:metrics}

We define the metrics used to characterize
individual architectures, compare independently
trained networks, and assess statistical differences.

\subsection{Reported Metrics and Notation}
\label{app:reported_metrics}

We denote the initial, best-checkpoint, and final neuron
populations by $N_0$, $N_{\mathrm{best}}$, and $N_f$,
respectively. Best-checkpoint statistics correspond to the
highest recorded test accuracy, while final statistics
correspond to the last training checkpoint. Cumulative
births and deaths count all neuron addition and removal
events. Architectural metrics and cross-run similarity
measures are defined in the following subsections.

\subsection{Individual Architecture Metrics}
\label{app:individual_metrics}

Each final architecture is characterized by neuron
count $N=N_f$, recurrent edge count $E_f$,
connectivity density $d$, and spectral radius $\rho$.
Let $W\in\mathbb{R}^{N\times N}$ denote the
recurrent weight matrix and
$A\in\{0,1\}^{N\times N}$ its binary connectivity
mask, where $A_{ij}=1$ indicates an existing
connection.

Connectivity density measures the fraction of
possible recurrent connections that are present:
\begin{equation}
d=\frac{1}{N^2}\sum_{i,j}A_{ij}
=\frac{E_f}{N^2}.
\label{eq:app_density}
\end{equation}

The spectral radius measures the largest absolute
eigenvalue of the recurrent weight matrix:
\begin{equation}
\rho=\max_{\lambda\in\sigma(W\odot A)}|\lambda|,
\label{eq:app_spectral_radius}
\end{equation}
where $\sigma(W\odot A)$ denotes the eigenvalue
spectrum of the masked recurrent matrix and
$\odot$ denotes elementwise multiplication.
It characterizes the spectral scale of the masked, unnormalized recurrent weight matrix, but does not directly characterize the stability of the degree-normalized recurrent dynamics used during inference.

\subsection{Cross-Run Architecture Similarity}
\label{app:comparative_metrics}

Independently trained networks are compared on a
common fixed probe set $\mathcal P$ of $M$ examples
($M=1000$ by default), randomly sampled from the
test partition. The same probe set is used for
all networks within each comparison. Comparisons are performed
both across random seeds at fixed $N_0$ and across
different initial population sizes.

\paragraph{Representational similarity.}
Let $H_a\in\mathbb{R}^{M\times N_a}$ and
$H_b\in\mathbb{R}^{M\times N_b}$ denote centered
hidden representations. Linear Centered Kernel
Alignment (CKA) is
\begin{equation}
\operatorname{CKA}(H_a,H_b)=
\frac{\|H_a^\top H_b\|_F^2}
{\|H_a^\top H_a\|_F\|H_b^\top H_b\|_F}.
\label{eq:app_cka}
\end{equation}
CKA permits comparison of representations with
different neuron counts
\citep{kornblith2019similarity}.

\paragraph{Neuron parameter-profile similarity.}
For each neuron $i$, we construct a parameter
profile
$p_i=[w_i^{\mathrm{in}},w_i^{\mathrm{out}},b_i]$
from its input weights, output weights, and
recurrent bias. Recurrent weights are excluded
because their coordinates depend on neuron ordering.

For neurons $i$ and $j$ from networks $a$ and $b$,
respectively, pairwise cosine similarity is
\begin{equation}
S_{ij}=
\frac{(p_i^{(a)})^\top p_j^{(b)}}
{\|p_i^{(a)}\|_2\|p_j^{(b)}\|_2}.
\end{equation}
The Hungarian algorithm \citep{kuhn1955hungarian}
determines the maximum-similarity one-to-one
matching $\pi^*$ between neurons of two networks.
We report
\begin{equation}
S_{\mathrm{Hungarian}}(a,b)=
\frac{1}{\min(N_a,N_b)}
\sum_{(i,j)\in\pi^*}S_{ij}.
\label{eq:app_hungarian}
\end{equation}
The matched-neuron fraction
$\min(N_a,N_b)/\max(N_a,N_b)$ is retained as a
diagnostic of population-size differences.

\paragraph{Functional similarity.}
Prediction agreement is the fraction of probe
examples assigned the same class:
\begin{equation}
S_{\mathrm{agree}}(a,b)=
\frac{1}{M}\sum_{x\in\mathcal P}
\mathbb{I}[\hat y_a(x)=\hat y_b(x)],
\label{eq:app_agreement}
\end{equation}
where $\mathbb{I}[\cdot]$ is the indicator function
and $\hat y_a(x)$ and $\hat y_b(x)$ are the predicted
classes of networks $a$ and $b$, respectively.

\subsection{Statistical Analysis}
\label{app:statistical_analysis}

For each experimental condition and initial network
size $N_0$, metrics are summarized across random
seeds using the mean and sample standard deviation.
Standard errors and coefficients of variation are
also computed.

Dependence of final neuron count, recurrent edge
count, connectivity density, and spectral radius
on $N_0$ is assessed using Welch's one-way ANOVA
\citep{welch1951comparison}, allowing unequal
group variances. Conventional ANOVA effect sizes
$\eta^2$ and $\omega^2$, computed from unweighted
sums of squares, are reported descriptively.

Because pairwise similarities share trained models,
they are not treated as independent observations.
For each model $a$, we compute a model-level
cross-$N_0$ score
\begin{equation}
\bar S_a^{\mathrm{cross}}=
\frac{1}{|\mathcal C_a|}
\sum_{b\in\mathcal C_a}S(a,b),
\qquad
\mathcal C_a=\{b:N_{0,b}\neq N_{0,a}\},
\label{eq:app_model_level}
\end{equation}
where comparisons are restricted to the same
experimental condition. Dependence of these scores
on source $N_0$ is examined using Welch's ANOVA.

For hyperparameter ablations, cross-setting
similarities compare models with different
parameter values at fixed $N_0$ and matched seeds.
For each source model, similarity is averaged
over the other settings within the same ablation.
Reported means and standard deviations are
computed across seeds.

Holm's procedure \citep{holm1979simple} corrects
$p$-values separately across the four individual
architecture metrics ($N_f$, $E_f$, $d$, and $\rho$)
and the three comparative metrics (CKA, Hungarian
similarity, and prediction agreement), within each
condition. Model-level similarity tests are
interpreted as exploratory because shared comparison
partners can introduce residual dependence.

\section{Hyperparameter Sensitivity Analysis}
\label{app:ablation_protocol}

We conduct systematic hyperparameter ablations on
Fashion-MNIST to characterize the sensitivity,
stability, and robustness of the proposed developmental
mechanism. Within each ablation, all parameters other than
the parameter under investigation are held fixed.
Structural-event limits may differ between
ablation types, as specified by their configurations. Each configuration
is evaluated over five random seeds, $\{0,1,2,3,4\}$.
Generalization experiments on CIFAR-10 and AG News
are presented separately in the main text.

\paragraph{Initial network size.}
We vary the initial neuron population over
$N_0 \in \{20,40,80,160\}$ on Fashion-MNIST.
This experiment examines the dependence of developed
network size and predictive performance on initialization.

\paragraph{Recurrent connectivity.}
We jointly vary the initial population
$N_0 \in \{10,20,40,80,160\}$ and recurrent
connectivity parameter
$\beta \in \{0.1,0.2,0.4,0.8\}$.
This experiment evaluates how initial connectivity
influences architectural development and stability.

\paragraph{Structural adaptation rate.}
We vary the structural rate
$r \in \{1,2.5,5,10\}$
for $N_0 \in \{20,40,80,160\}$.
The rate specifies the maximum number of births and
deaths permitted per structural check.
For $r=2.5$, the limits alternate between
two and three across consecutive checks.

\paragraph{Warm-up duration.}
At $N_0=80$, we vary the number of training examples
processed before structural adaptation is enabled:
\[
T_{\mathrm{warm}} \in
\{50\mathrm{k},100\mathrm{k},200\mathrm{k},400\mathrm{k},800\mathrm{k}\}.
\]
This experiment evaluates the influence of delayed
structural adaptation on developmental trajectories
and final performance.

\paragraph{Birth threshold.}
At $N_0=20$, we vary the birth threshold over
$\{0.15,0.25,0.35,0.45\}$.
This experiment measures the sensitivity of neuronal
growth, final architecture, and predictive performance
to the criterion governing neuron addition.

\paragraph{Death criteria.}
At $N_0=20$, we independently vary two pruning
parameters. The death-Flux fraction takes values
$\{0.005,\allowbreak 0.05,\allowbreak
0.5,\allowbreak 0.95\}$, while the
death-activity fraction takes values
$\{0.05,\allowbreak 0.10,\allowbreak
0.20,\allowbreak 0.40\}$.
These experiments assess the sensitivity of pruning
dynamics and architectural stability to the
criteria governing neuron removal.

\paragraph{Evaluation.}
We record predictive accuracy, neuron population,
cumulative births and deaths, and structural events
throughout training. Final architectures are
characterized by recurrent edge count, connectivity
density, and spectral radius.
Cross-setting similarity is quantified using centered
kernel alignment (CKA) and prediction agreement
between models trained under different hyperparameter
values at fixed $N_0$ and matched seeds.
For each source model, similarity is averaged across
the other settings within the same ablation.
All results are reported as mean $\pm$ standard
deviation across five seeds.

\paragraph{Default configuration.}
Unless otherwise specified, experiments use the
following configuration:

\begin{itemize}
    \setlength{\itemsep}{1pt}
    \setlength{\parskip}{0pt}

    \item \textbf{Training:}
    100 epochs, batch size 128, Adam optimizer,
    learning rate $10^{-3}$, and weight decay $10^{-4}$.

    \item \textbf{Network architecture:}
    10 relaxation steps, recurrent connectivity
    $\beta=0.2$, and no self-connections.
    The neuron population is constrained to
    $10 \leq N \leq 300$.

\item \textbf{Neurogenesis:}
The default birth threshold is $0.25$.
The birth persistence interval is 10,000 training
examples, and the neuron-splitting noise standard
deviation is $0.01$. The birth fraction is $0.20$.
The default maximum number of births per structural
check is $5$.

\item \textbf{Pruning:} The default death-Flux and death-activity
quantile fractions are $0.01$ and $0.10$,
respectively, and the absolute death-Flux
threshold is $0.13$.
The death persistence interval is 10,000 training
examples. The default maximum number of deaths
per structural check is $5$.

    \item \textbf{Structural dynamics:}
    Flux and activity exponential moving-average
    decay factors are both $0.90$.
    Structural adaptation begins after
    100,000 training examples, with checks every
    10,000 examples and a cooldown of
    10,000 examples.
\end{itemize}

\section{Fashion-MNIST Ablations}
\label{sec:fm_ablations}

\begin{table*}[t]
\centering
\caption{Unified FlowGenesis hyperparameter ablations on Fashion-MNIST
(five seeds per setting). Values are mean $\pm$ standard deviation.
Section headings indicate the ablation type and initial neuron count $N_0$.}
\label{tab:all_ablations}
\scriptsize
\setlength{\tabcolsep}{2pt}
\renewcommand{\arraystretch}{1.12}
\resizebox{\textwidth}{!}{%
\begin{tabular}{@{}l cc cc cc ccc cc@{}}
\toprule
& \multicolumn{2}{c}{Network size}
& \multicolumn{2}{c}{Performance}
& \multicolumn{2}{c}{Structural dynamics}
& \multicolumn{3}{c}{Final structure}
& \multicolumn{2}{c}{Cross-setting similarity} \\
\cmidrule(lr){2-3}
\cmidrule(lr){4-5}
\cmidrule(lr){6-7}
\cmidrule(lr){8-10}
\cmidrule(lr){11-12}
Value
& $N_{\mathrm{best}}$ & $N_f$
& Best acc. & Final acc.
& Births & Pruned
& $E_f$ & Density & $\rho$
& CKA & Agr. \\
\midrule

\multicolumn{12}{c}{\textbf{Connectivity ($N_0=20$)}} \\
\midrule
0.1
& $54.6\pm9.7$ & $53.4\pm7.8$
& $87.42\pm0.29$ & $86.89\pm0.41$
& $621.4\pm174.6$ & $588.0\pm168.3$
& $973.6\pm577.0$ & $0.320\pm0.085$ & $0.568\pm0.134$
& $0.894\pm0.015$ & $92.60\pm0.61$ \\

0.2
& $55.8\pm9.4$ & $53.2\pm5.9$
& $87.51\pm0.19$ & $87.01\pm0.38$
& $632.0\pm40.3$ & $598.8\pm37.4$
& $1017.4\pm153.1$ & $0.362\pm0.041$ & $0.537\pm0.067$
& $0.896\pm0.026$ & $92.85\pm0.38$ \\

0.4
& $49.8\pm5.7$ & $55.0\pm7.3$
& $87.53\pm0.10$ & $86.79\pm0.38$
& $624.0\pm161.0$ & $589.0\pm167.3$
& $1608.6\pm531.4$ & $0.520\pm0.038$ & $0.458\pm0.090$
& $0.891\pm0.032$ & $92.73\pm0.39$ \\

0.8
& $51.8\pm5.9$ & $54.6\pm10.3$
& $87.53\pm0.39$ & $86.79\pm0.55$
& $620.4\pm90.4$ & $585.8\pm90.4$
& $2594.6\pm985.2$ & $0.851\pm0.044$ & $0.327\pm0.029$
& $0.902\pm0.021$ & $92.66\pm0.38$ \\

\midrule
\multicolumn{12}{c}{\textbf{Warm-up duration ($N_0=80$)}} \\
\midrule
50k
& $53.6\pm2.9$ & $51.6\pm5.2$
& $87.49\pm0.31$ & $87.05\pm0.64$
& $474.8\pm60.4$ & $503.2\pm61.9$
& $891.8\pm305.4$ & $0.332\pm0.106$ & $0.528\pm0.135$
& $0.883\pm0.056$ & $92.93\pm0.42$ \\

100k
& $51.6\pm6.3$ & $52.6\pm5.5$
& $87.42\pm0.27$ & $86.95\pm0.51$
& $461.6\pm63.1$ & $489.0\pm64.7$
& $906.8\pm131.7$ & $0.330\pm0.039$ & $0.520\pm0.079$
& $0.891\pm0.032$ & $92.74\pm0.81$ \\

200k
& $51.4\pm7.0$ & $49.8\pm7.5$
& $87.61\pm0.18$ & $86.96\pm0.19$
& $395.0\pm61.9$ & $425.2\pm59.2$
& $790.0\pm156.2$ & $0.323\pm0.054$ & $0.532\pm0.098$
& $0.896\pm0.025$ & $92.94\pm0.42$ \\

400k
& $53.4\pm8.8$ & $55.2\pm8.5$
& $87.66\pm0.19$ & $86.97\pm0.15$
& $367.6\pm70.6$ & $392.4\pm65.7$
& $926.8\pm342.1$ & $0.295\pm0.069$ & $0.570\pm0.166$
& $0.890\pm0.019$ & $92.92\pm0.85$ \\

800k
& $75.4\pm10.3$ & $52.0\pm5.6$
& $87.82\pm0.22$ & $86.80\pm0.49$
& $339.6\pm55.2$ & $367.6\pm51.2$
& $737.4\pm161.3$ & $0.273\pm0.053$ & $0.578\pm0.028$
& $0.898\pm0.020$ & $92.93\pm0.48$ \\

\midrule
\multicolumn{12}{c}{\textbf{Structural adaptation rate ($N_0=160$)}} \\
\midrule
1
& $102.8\pm16.5$ & $47.0\pm3.3$
& $88.25\pm0.22$ & $86.68\pm0.85$
& $179.2\pm27.2$ & $292.2\pm28.3$
& $593.8\pm73.3$ & $0.272\pm0.047$ & $0.575\pm0.094$
& $0.891\pm0.030$ & $91.97\pm0.45$ \\

2.5
& $50.0\pm5.2$ & $55.8\pm8.9$
& $87.68\pm0.18$ & $87.07\pm0.25$
& $272.8\pm47.2$ & $377.0\pm45.3$
& $684.0\pm212.2$ & $0.216\pm0.035$ & $0.761\pm0.176$
& $0.879\pm0.039$ & $92.61\pm0.51$ \\

5
& $51.8\pm6.6$ & $51.4\pm4.2$
& $87.64\pm0.14$ & $86.66\pm0.29$
& $295.8\pm29.3$ & $404.4\pm31.4$
& $631.2\pm168.5$ & $0.235\pm0.031$ & $0.699\pm0.149$
& $0.886\pm0.029$ & $92.45\pm0.71$ \\

10
& $50.8\pm4.2$ & $54.4\pm6.8$
& $87.54\pm0.16$ & $86.51\pm0.22$
& $337.6\pm49.3$ & $443.2\pm51.2$
& $681.4\pm145.1$ & $0.233\pm0.044$ & $0.510\pm0.053$
& $0.884\pm0.025$ & $92.27\pm0.97$ \\

\midrule
\multicolumn{12}{c}{\textbf{Death-flux fraction ($N_0=20$)}} \\
\midrule
0.005
& $105.4\pm17.5$ & $54.8\pm18.1$
& $88.71\pm0.16$ & $86.91\pm0.44$
& $406.4\pm30.5$ & $371.6\pm41.4$
& $1524.6\pm1392.6$ & $0.425\pm0.086$ & $0.574\pm0.106$
& $0.902\pm0.018$ & $92.07\pm0.62$ \\

0.05
& $51.6\pm6.4$ & $54.4\pm5.9$
& $87.60\pm0.14$ & $86.81\pm0.35$
& $656.4\pm63.1$ & $622.0\pm60.4$
& $1193.4\pm232.9$ & $0.406\pm0.076$ & $0.686\pm0.303$
& $0.911\pm0.013$ & $92.92\pm0.58$ \\

0.5
& $55.4\pm5.3$ & $53.2\pm10.0$
& $87.41\pm0.16$ & $86.80\pm0.58$
& $716.4\pm162.5$ & $683.2\pm154.5$
& $1113.6\pm448.4$ & $0.378\pm0.020$ & $0.649\pm0.086$
& $0.921\pm0.049$ & $94.35\pm1.62$ \\

0.95
& $55.6\pm5.0$ & $52.0\pm10.0$
& $87.43\pm0.18$ & $87.02\pm0.34$
& $716.6\pm162.5$ & $684.6\pm154.3$
& $1086.0\pm446.7$ & $0.386\pm0.028$ & $0.629\pm0.104$
& $0.930\pm0.030$ & $94.45\pm1.46$ \\

\midrule
\multicolumn{12}{c}{\textbf{Birth threshold ($N_0=20$)}} \\
\midrule
0.15
& $168.8\pm13.7$ & $154.6\pm19.8$
& $88.06\pm0.19$ & $87.42\pm0.25$
& $5959.2\pm144.6$ & $5824.6\pm128.1$
& $7914.2\pm1037.0$ & $0.341\pm0.084$ & $0.488\pm0.150$
& $0.863\pm0.034$ & $92.37\pm0.62$ \\

0.25 (default)
& $52.4\pm6.4$ & $56.0\pm8.1$
& $87.51\pm0.22$ & $86.90\pm0.19$
& $712.6\pm104.7$ & $676.6\pm108.5$
& $1194.6\pm353.3$ & $0.376\pm0.043$ & $0.669\pm0.100$
& $0.892\pm0.026$ & $92.36\pm0.48$ \\

0.35
& $32.4\pm4.4$ & $28.2\pm1.6$
& $86.99\pm0.18$ & $86.49\pm0.28$
& $127.2\pm59.9$ & $119.0\pm58.8$
& $276.0\pm82.8$ & $0.350\pm0.120$ & $0.603\pm0.091$
& $0.892\pm0.024$ & $92.29\pm0.31$ \\

0.45
& $24.2\pm3.7$ & $20.4\pm1.3$
& $86.89\pm0.20$ & $86.15\pm0.47$
& $37.2\pm15.9$ & $36.8\pm15.2$
& $118.2\pm12.9$ & $0.286\pm0.038$ & $0.622\pm0.124$
& $0.887\pm0.027$ & $92.02\pm0.49$ \\

\midrule
\multicolumn{12}{c}{\textbf{Death-activity fraction ($N_0=20$)}} \\
\midrule
0.05
& $50.4\pm2.1$ & $49.0\pm6.7$
& $87.60\pm0.14$ & $86.99\pm0.40$
& $625.8\pm106.5$ & $596.8\pm106.3$
& $891.0\pm351.8$ & $0.359\pm0.055$ & $0.630\pm0.061$
& $0.904\pm0.021$ & $92.99\pm0.67$ \\

0.10 (default)
& $52.4\pm6.4$ & $56.0\pm8.1$
& $87.51\pm0.22$ & $86.90\pm0.19$
& $712.6\pm104.7$ & $676.6\pm108.5$
& $1194.6\pm353.3$ & $0.376\pm0.043$ & $0.669\pm0.100$
& $0.894\pm0.029$ & $92.62\pm0.66$ \\

0.20
& $56.2\pm6.1$ & $60.0\pm12.2$
& $87.42\pm0.21$ & $86.64\pm0.28$
& $868.6\pm151.7$ & $828.6\pm154.2$
& $1263.6\pm487.8$ & $0.347\pm0.056$ & $0.621\pm0.149$
& $0.897\pm0.031$ & $92.81\pm0.92$ \\

0.40
& $52.6\pm7.0$ & $59.6\pm9.8$
& $87.35\pm0.10$ & $86.61\pm0.38$
& $1015.8\pm102.3$ & $976.2\pm99.5$
& $1215.8\pm559.7$ & $0.324\pm0.060$ & $0.669\pm0.133$
& $0.879\pm0.045$ & $93.07\pm0.77$ \\

\bottomrule
\end{tabular}%
}
\vspace{3pt}

\begin{minipage}{\textwidth}
\footnotesize
\textit{Notes:}
$N_{\mathrm{best}}$ and $N_f$ denote neuron counts at best
accuracy and training end; $E_f$, density, and $\rho$ denote
final recurrent edges, connectivity density, and spectral radius.
Accuracies and prediction agreement are percentages.
CKA measures representational similarity.
Cross-setting similarities compare different hyperparameter values
at fixed $N_0$ and matched seeds; each model is averaged over
other settings, with SD across five seeds.
All ablations include four settings except warm-up (five);
death-flux includes seven, with four shown.
Initial $N_0$ varies across ablation types, precluding
controlled cross-type comparisons.
\end{minipage}

\end{table*}
We evaluate the sensitivity of FlowGenesis to initial
connectivity, structural warm-up, adaptation rate,
and neuron birth and death criteria on Fashion-MNIST.
Each configuration is evaluated across five seeds.
Table~\ref{tab:all_ablations} summarizes final
architecture, predictive performance, structural
turnover, and cross-setting similarity.

\begin{figure*}[!t]
\centering

\begin{subfigure}[t]{0.49\textwidth}
    \centering
    \includegraphics[
        width=\linewidth,
        height=0.19\textheight,
        keepaspectratio
    ]{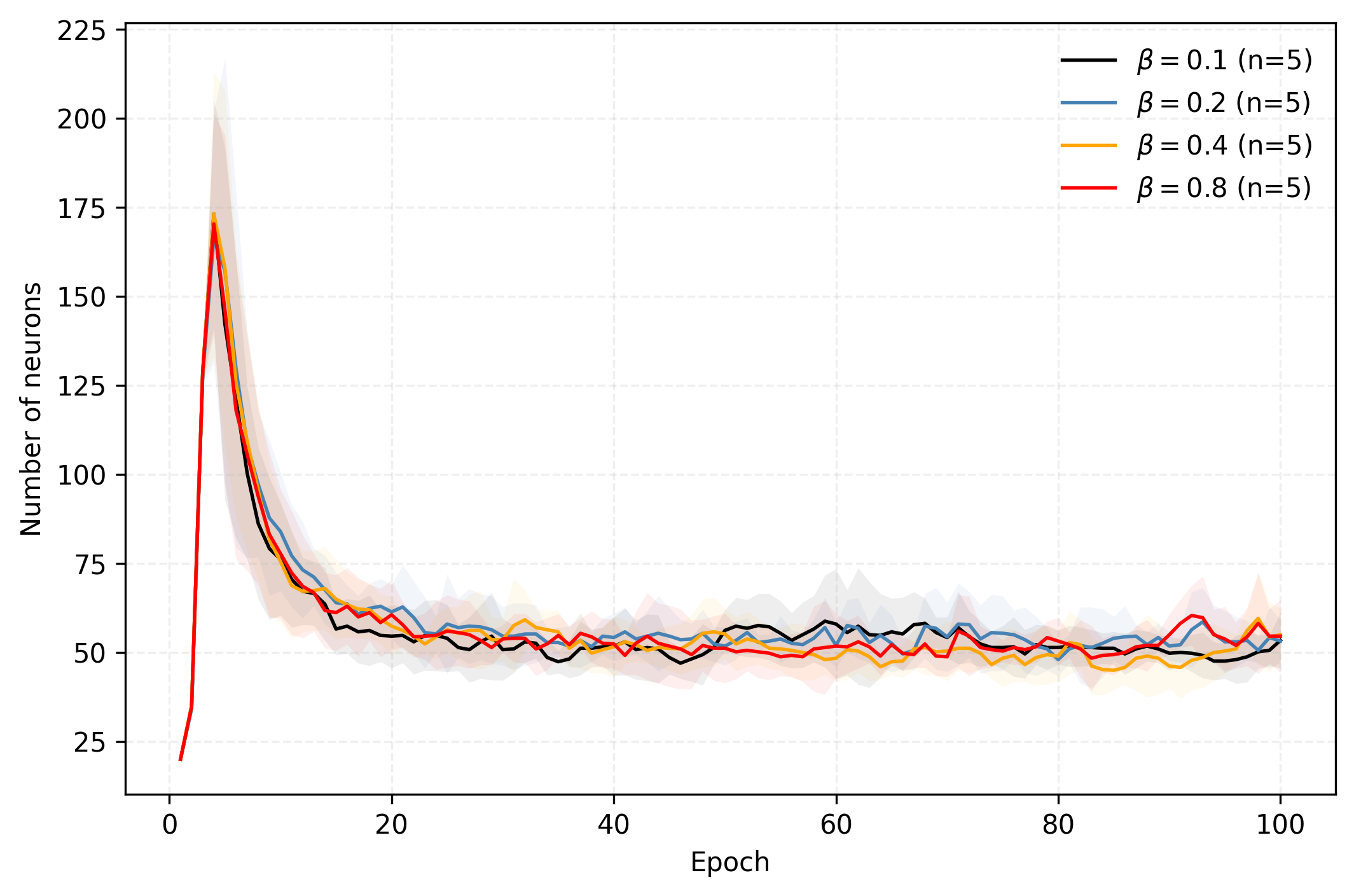}
    \caption{Connectivity: neuron count.}
\end{subfigure}
\hfill
\begin{subfigure}[t]{0.49\textwidth}
    \centering
    \includegraphics[
        width=\linewidth,
        height=0.19\textheight,
        keepaspectratio
    ]{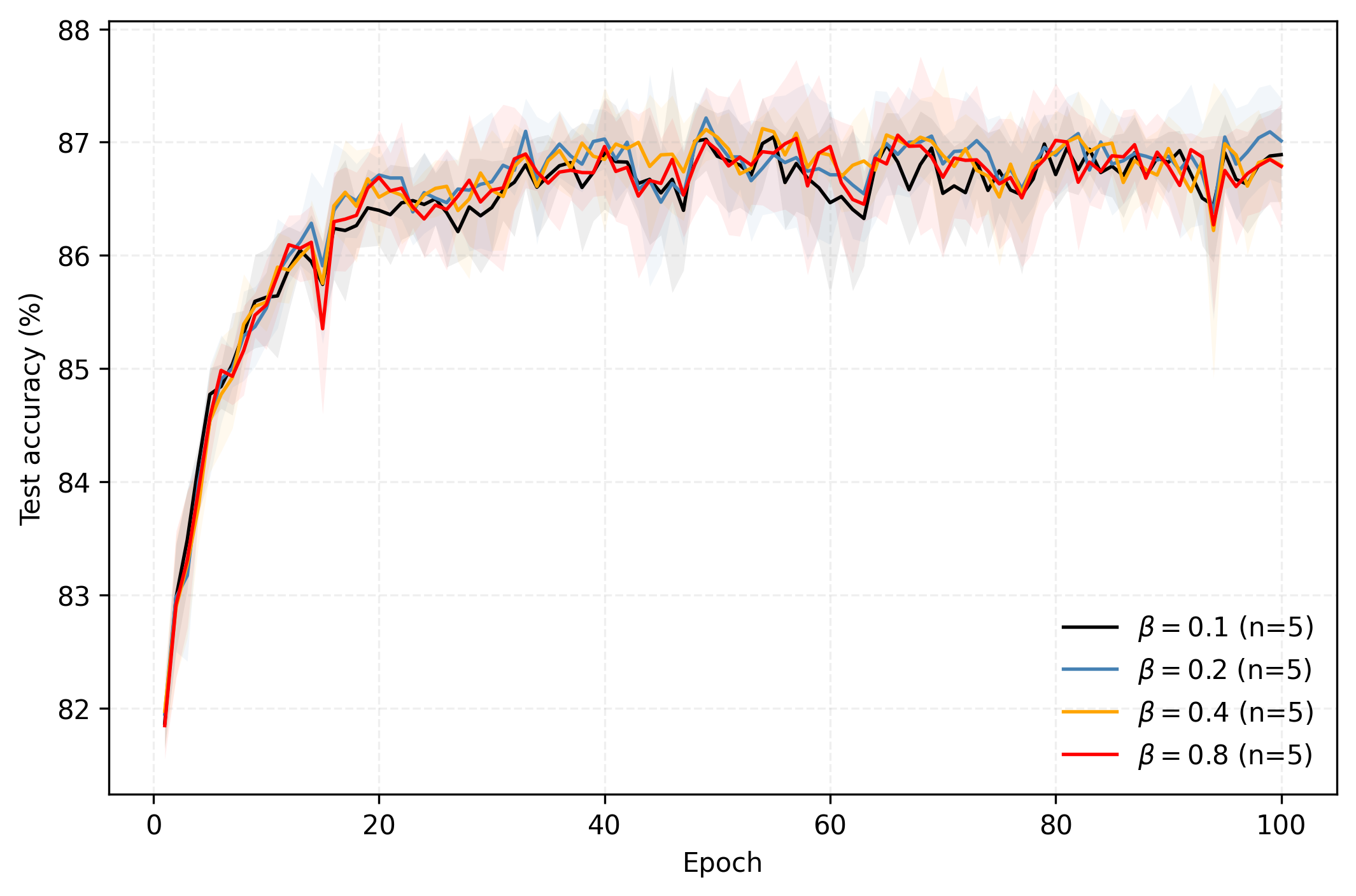}
    \caption{Connectivity: test accuracy.}
\end{subfigure}

\vspace{0.2em}

\begin{subfigure}[t]{0.49\textwidth}
    \centering
    \includegraphics[
        width=\linewidth,
        height=0.19\textheight,
        keepaspectratio
    ]{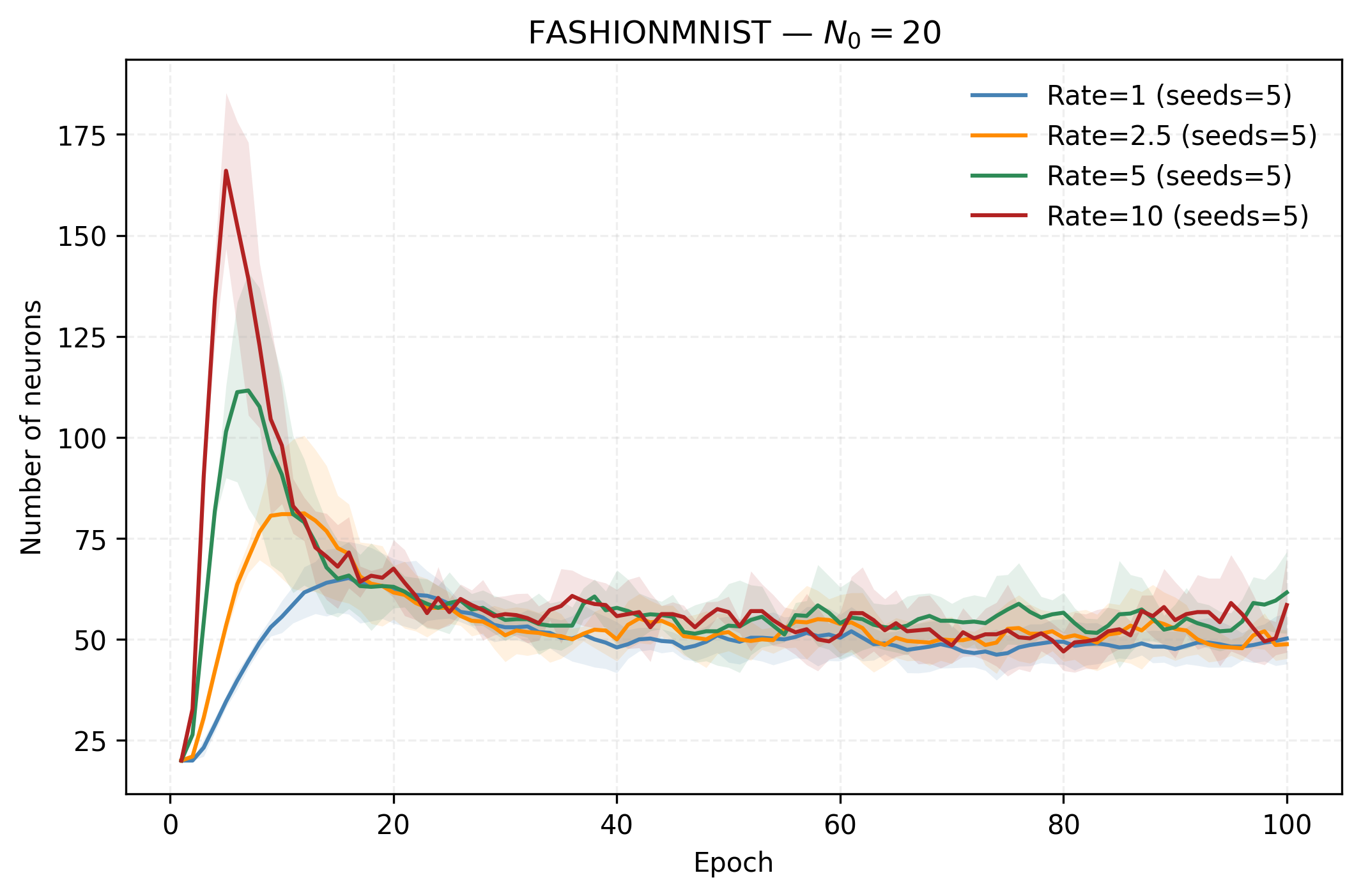}
    \caption{Adaptation rate: neuron count.}
\end{subfigure}
\hfill
\begin{subfigure}[t]{0.49\textwidth}
    \centering
    \includegraphics[
        width=\linewidth,
        height=0.19\textheight,
        keepaspectratio
    ]{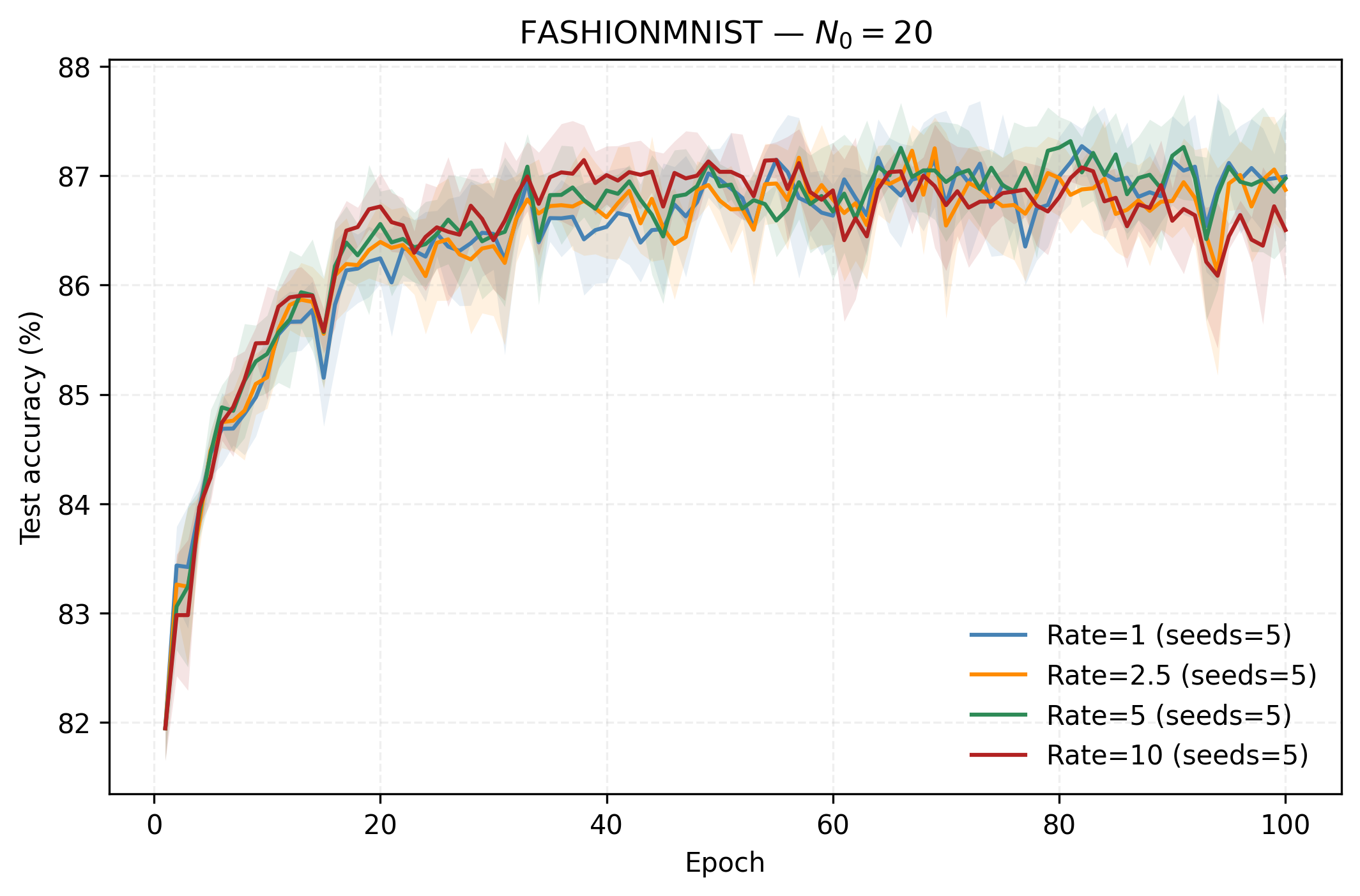}
    \caption{Adaptation rate: test accuracy.}
\end{subfigure}

\vspace{0.2em}

\begin{subfigure}[t]{0.49\textwidth}
    \centering
    \includegraphics[
        width=\linewidth,
        height=0.19\textheight,
        keepaspectratio
    ]{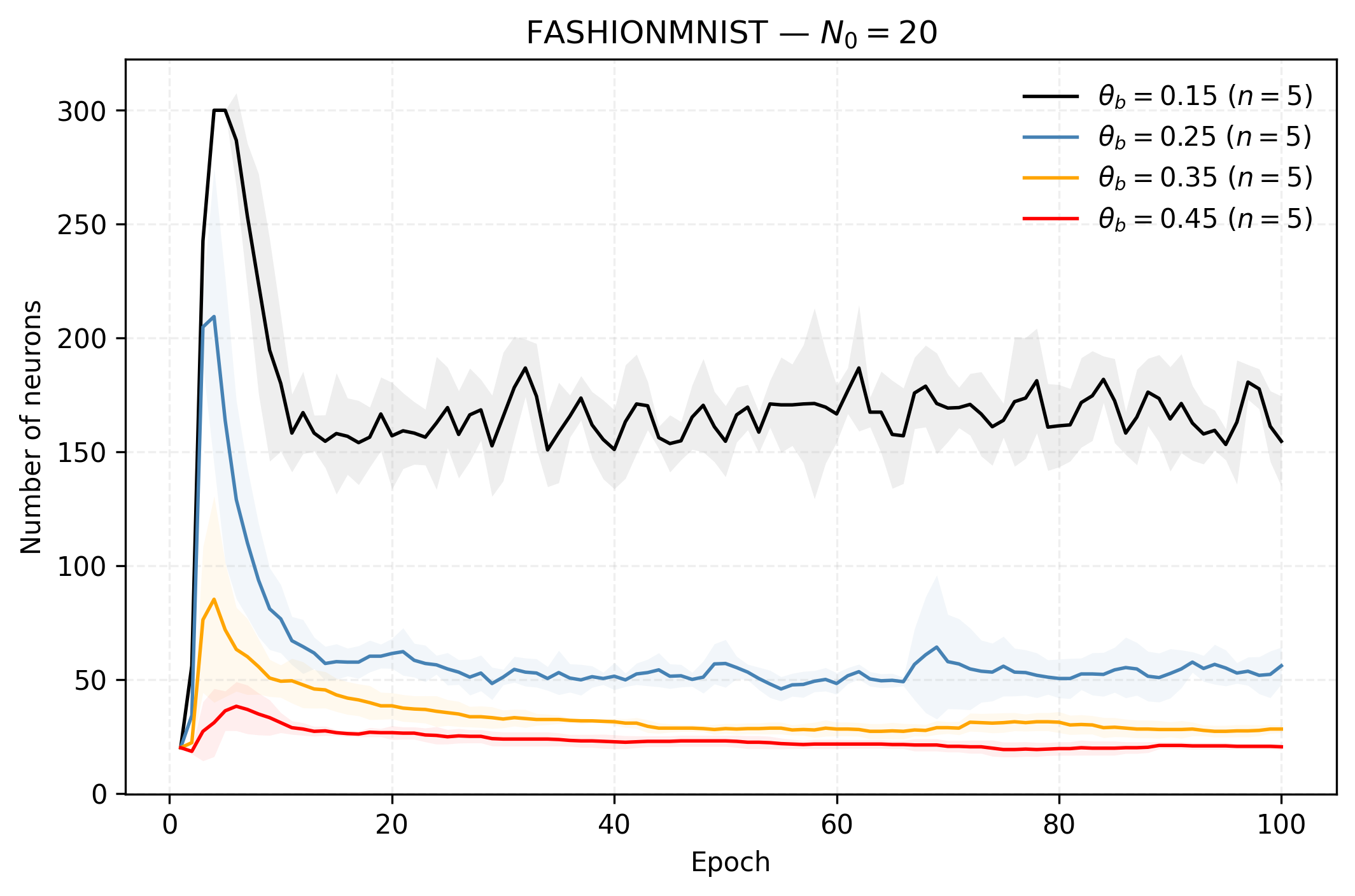}
    \caption{Birth threshold: neuron count.}
\end{subfigure}
\hfill
\begin{subfigure}[t]{0.49\textwidth}
    \centering
    \includegraphics[
        width=\linewidth,
        height=0.19\textheight,
        keepaspectratio
    ]{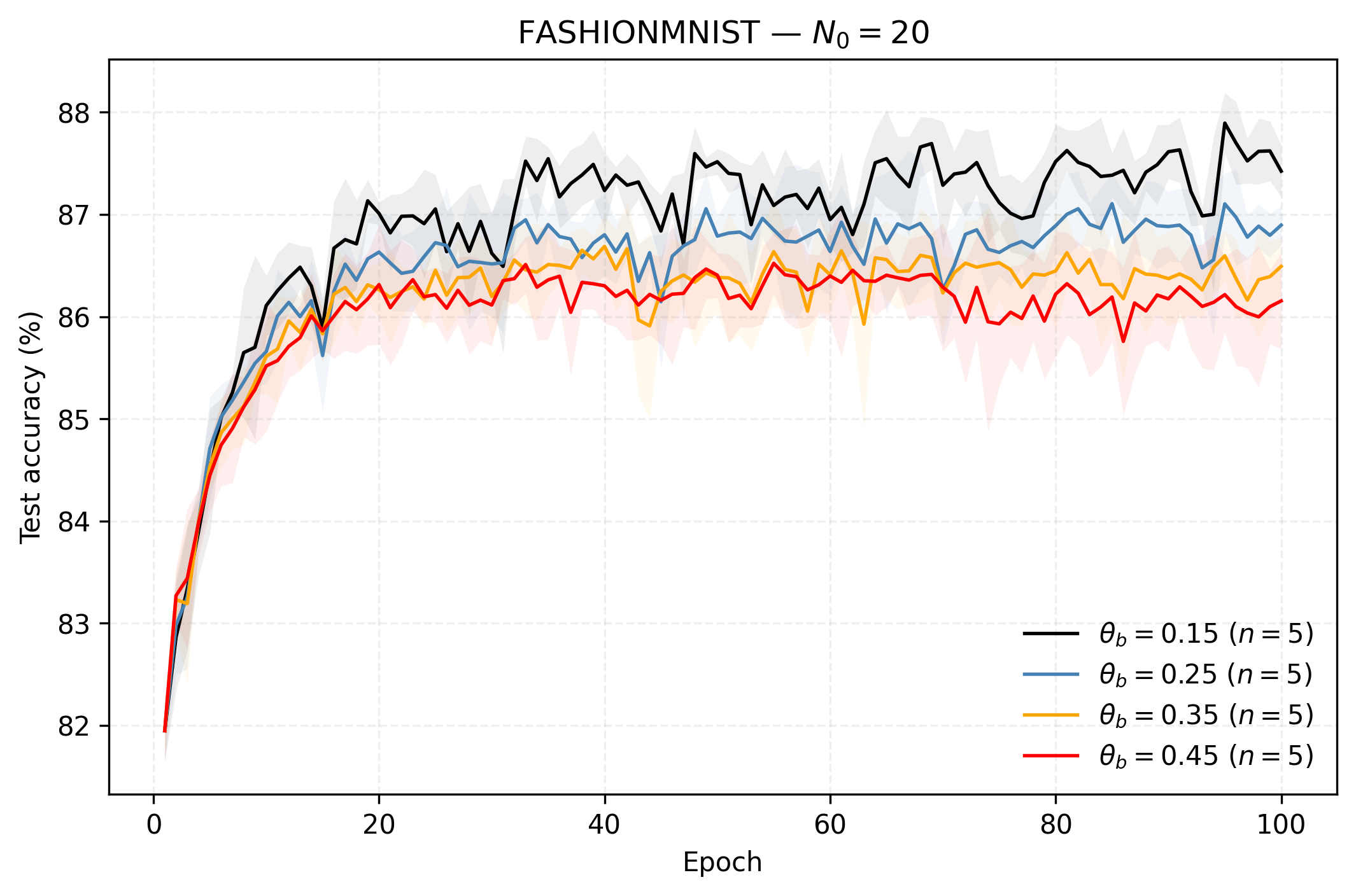}
    \caption{Birth threshold: test accuracy.}
\end{subfigure}

\caption{
Fashion-MNIST structural ablations ($N_0=20$).
(a,b) Initial connectivity
$\beta\in\{0.1,0.2,0.4,0.8\}$;
(c,d) adaptation rate
$r\in\{1,2.5,5,10\}$;
(e,f) birth threshold
$\theta_{\mathrm{birth}}\in\{0.15,0.25,0.35,0.45\}$.
Left: neuron count; right: test accuracy.
Curves show mean $\pm$ standard deviation
across five seeds.
Final populations are relatively insensitive
to connectivity and adaptation rate,
but strongly depend on the birth threshold.
}
\label{fig:fm_ablation_dynamics}

\end{figure*}
\paragraph{Initial connectivity.}
Varying the initial connectivity probability
$\beta$ from $0.1$ to $0.8$ produces similar
final populations ($53.2$--$55.0$ neurons)
and best accuracies ($87.42$--$87.53\%$).
Despite substantial differences in final edge
counts and density, the networks exhibit
comparable representations (CKA $0.891$--$0.902$).
Fig.~\ref{fig:fm_ablation_dynamics}(a,b) shows
convergence in population size and predictive
performance across connectivity settings.

\paragraph{Structural warm-up.}
Increasing $T_{\mathrm{warm}}$ from 50k to 800k
processed examples has limited effect on final
network size ($49.8$--$55.2$ neurons) or best
accuracy ($87.42$--$87.82\%$).
Longer warm-up reduces cumulative structural
events, partly because fewer adaptation
opportunities remain. Thus, the final population
is relatively insensitive to the onset of
structural adaptation
(Table~\ref{tab:all_ablations}).

\paragraph{Structural adaptation rate.}
Across $r\in\{1,2.5,5,10\}$, the final
population remains within $47.0$--$55.8$
neurons in the $N_0=160$ experiments, while
best accuracy ranges from $87.54\%$ to
$88.25\%$. Higher rates generally increase
structural turnover. The separate $N_0=20$
trajectories in Fig.~\ref{fig:fm_ablation_dynamics}(c,d)
illustrate differences in transient expansion
despite broadly comparable late-training
population sizes and accuracy.

\paragraph{Birth threshold.}
The birth threshold has the strongest effect
on final capacity. Increasing
$\theta_{\mathrm{birth}}$ from $0.15$ to $0.45$
reduces the final population from
$154.6\pm19.8$ to $20.4\pm1.3$ neurons
and cumulative births from approximately
5960 to 37. Best accuracy decreases from
$88.06\%$ to $86.89\%$.
Fig.~\ref{fig:fm_ablation_dynamics}(e,f)
shows that permissive birth thresholds
produce substantially larger architectures.
Unlike the preceding controls, this parameter
strongly influences the final population
regime.

\paragraph{Death-Flux fraction.}
Varying the pruning Flux fraction from
$0.005$ to $0.95$ changes structural turnover
but leaves final populations within
$52.0$--$54.8$ neurons for the settings
reported in Table~\ref{tab:all_ablations}.
Best accuracy ranges from $87.41\%$ to
$88.71\%$, while cross-setting CKA remains
high ($0.902$--$0.930$).
The most restrictive setting exhibits fewer
structural events and a larger population
at peak accuracy.

\paragraph{Death-activity fraction.}
Increasing the activity-based pruning fraction
from $0.05$ to $0.40$ increases cumulative
births and deaths, while final populations
remain between $49.0$ and $60.0$ neurons.
Best accuracy varies from $87.35\%$ to
$87.60\%$, and cross-setting CKA remains
between $0.879$ and $0.904$.
Thus, this parameter primarily affects
structural turnover rather than final
predictive performance.

\section{Flux-Guided Neurogenesis: Splitting and Stabilization}
\label{app:flux_neurogenesis}

We analyze three aspects of Flux-guided neurogenesis:
the effect of splitting on the represented function,
the accuracy of Flux-based structural selection, and
the conditions for subsequent stabilization.

\subsection{Local dynamics.}

\label{app:local_splitting}

\paragraph{Proposition 1 (Local functional
stability of neuron splitting).}
Consider the FlowGenesis recurrent update
\begin{equation}
h^{(s+1)}
=
(1-\gamma)h^{(s)}
+
\gamma\operatorname{ReLU}
\left(u+Wh^{(s)}+b\right),
\end{equation}
with $h^{(0)}=0$ and output
$f(x)=W_{\mathrm{out}}h^{(S)}+b_{\mathrm{out}}$.
Here $W$ is the effective,
degree-normalized recurrent matrix.

Suppose one neuron is split, increasing
the population from $N$ to $N+1$.
Let $P\in\{0,1\}^{(N+1)\times N}$
denote the duplication matrix that copies
the split neuron's coordinate while
preserving all others. Let $W\in\mathbb{R}^{N\times N}$,
$u,b\in\mathbb{R}^{N}$ denote the
effective recurrent matrix, input projection,
and bias of the original network.
Their counterparts in the expanded network
are denoted by
$W^+\in\mathbb{R}^{(N+1)\times(N+1)}$
and $u^+,b^+\in\mathbb{R}^{N+1}$.
Assume that, for a fixed input $x$,
$\|h^{(s)}\|\leq H$ for $s=0,\ldots,S$.
Define
\begin{equation}
\varepsilon
=
\|W^+P-PW\|H
+
\|u^+-Pu\|
+
\|b^+-Pb\|,
\end{equation}
and
\begin{equation}
L=1+\gamma(\|W^+\|-1).
\end{equation}
Here, $\varepsilon$ bounds the disturbance
introduced by splitting into the recurrent
dynamics, while $L$ bounds the amplification
of existing perturbations during each
relaxation step.

Let
$f^+(x)=W_{\mathrm{out}}^+h_+^{(S)}
+b_{\mathrm{out}}^+$
denote the output of the expanded network,
where $h_+^{(S)}$ is its hidden state after
$S$ relaxation steps.

If the expanded output layer satisfies
\begin{equation}
W_{\mathrm{out}}^+P=W_{\mathrm{out}},
\qquad
b_{\mathrm{out}}^+=b_{\mathrm{out}},
\end{equation}
then
\begin{equation}
\boxed{
\|f^+(x)-f(x)\|
\leq
\gamma\|W_{\mathrm{out}}^+\|
\varepsilon
\sum_{j=0}^{S-1}L^j.
}
\label{eq:fg_splitting_stability}
\end{equation}
The bound follows from a standard
Lipschitz-based perturbation argument
applied to the recurrent dynamics.

\emph{Proof.}
Let $h_+^{(s)}$ denote the expanded
network's hidden state and define
\[
\delta_s=\|h_+^{(s)}-Ph^{(s)}\|.
\]
Since both networks start from zero,
$\delta_0=0$.

Because ReLU is 1-Lipschitz and commutes
with coordinate duplication, the difference
between the expanded and duplicated original
updates is bounded by
\begin{align}
\delta_{s+1}
&\leq
(1-\gamma)\delta_s
+
\gamma\left(
\|W^+\|\delta_s+\varepsilon
\right)
\nonumber\\
&=
L\delta_s+\gamma\varepsilon.
\end{align}
Here, the first term represents the
propagation of the existing hidden-state
discrepancy, while $\varepsilon$ bounds
the additional mismatch introduced by
the expanded network's parameters.

Since $\delta_0=0$, successive application
of this inequality gives
\begin{align}
\delta_1
&\leq \gamma\varepsilon,
\\
\delta_2
&\leq \gamma\varepsilon(1+L).
\end{align}
Continuing recursively, we obtain
\begin{equation}
\delta_S
\leq
\gamma\varepsilon
\sum_{j=0}^{S-1}L^j.
\end{equation}

Output compatibility implies
\begin{equation}
f^+(x)-f(x)
=
W_{\mathrm{out}}^+
(h_+^{(S)}-Ph^{(S)}).
\end{equation}
Taking norms proves the result.
\hfill$\square$

Proposition~1 bounds the immediate
functional perturbation caused by neuron
splitting. In FlowGenesis, the parent's
output weights are divided equally
between parent and child, while the
output bias remains unchanged.  Particularly, Gaussian perturbations
of copied weights can alter the child's
response relative to its parent.
New recurrent connections introduce
additional interactions, while degree
normalization can change effective
recurrent weights as incoming connectivity
changes. These effects may produce
nonzero deviations from exact
duplication-compatible dynamics.

\paragraph{Neuron splitting and Representation Flux.}
Neuron splitting introduces additional trainable
degrees of freedom while aiming to preserve the
existing computation. FlowGenesis selects neurons for splitting using
Representation Flux rather than explicitly
estimating splitting curvature or optimization
benefit. Let $h_i(x;\theta_t)$ denote the
activation of neuron $i$ for input $x$, where
$\theta_t$ represents the network parameters
at training step $t$. The mean activation over
a minibatch $B_t$ of size $m$ is

\begin{equation}
\bar h_{i,t}
=
\frac{1}{m}
\sum_{x\in B_t}h_i(x;\theta_t).
\end{equation}

Representation Flux is the absolute difference
between consecutive minibatch mean activations:

\begin{equation}
\Phi_{i,t}
=
|\bar h_{i,t}-\bar h_{i,t-1}|.
\end{equation}

Unlike function-preserving network widening
\citep{chen2016net2net} and curvature-based
splitting \citep{wu2019splitting}, FlowGenesis
uses changes in neuronal activation as a
heuristic signal for allocating additional
capacity. To distinguish changes in expected activation
from minibatch sampling variability, let $P$
denote the data distribution and define

\begin{equation}
\mu_i(\theta)
=
\mathbb E_{x\sim P}[h_i(x;\theta)],
\qquad
\sigma_i^2(\theta)
=
\operatorname{Var}_{x\sim P}[h_i(x;\theta)].
\end{equation}

Here, $\mu_i(\theta)$ and $\sigma_i^2(\theta)$
are the population mean and variance of neuron
$i$'s activation. The observed minibatch mean
can be expressed as

\begin{equation}
\bar h_{i,t}
=
\mu_i(\theta_t)+\xi_{i,t},
\end{equation}

where $\xi_{i,t}$ is the sampling error due
to the finite minibatch.

\paragraph{Proposition 2 (Sampling-noise decomposition).}
Suppose that, conditional on fixed network
parameters $\theta_t$ and $\theta_{t-1}$,
the minibatches $B_t$ and $B_{t-1}$
are drawn independently from $P$,
with $m$ independent samples per minibatch.

Then, conditional on these parameters, the
expected squared Representation Flux satisfies

\begin{align}
\mathbb E[
\Phi_{i,t}^2
\mid\theta_t,\theta_{t-1}]
&=
\left[
\mu_i(\theta_t)-\mu_i(\theta_{t-1})
\right]^2
\nonumber\\
&\quad+
\frac{
\sigma_i^2(\theta_t)+
\sigma_i^2(\theta_{t-1})
}{m}.
\label{eq:fg_noise}
\end{align}

\emph{Proof.}
Define the signed difference between consecutive
minibatch mean activations as

\begin{equation}
D_{i,t}
=
\bar h_{i,t}-\bar h_{i,t-1},
\end{equation}

so that $\Phi_{i,t}^2=D_{i,t}^2$.
Substituting the decomposition of the minibatch
means gives

\begin{align}
D_{i,t}
&=
\mu_i(\theta_t)-\mu_i(\theta_{t-1})
\nonumber\\
&\quad+
\xi_{i,t}-\xi_{i,t-1}.
\end{align}

Under the stated sampling assumptions, the
errors have zero conditional means and

\begin{equation}
\operatorname{Var}(
\xi_{i,t}\mid\theta_t,\theta_{t-1})
=
\frac{\sigma_i^2(\theta_t)}{m},
\end{equation}

with the analogous expression for
$\xi_{i,t-1}$. Independence of the minibatches
implies that their sampling errors have zero
conditional covariance. Therefore,

\begin{align}
\mathbb E[
D_{i,t}^2\mid\theta_t,\theta_{t-1}]
&=
\left[
\mu_i(\theta_t)-\mu_i(\theta_{t-1})
\right]^2
\nonumber\\
&\quad+
\frac{
\sigma_i^2(\theta_t)+
\sigma_i^2(\theta_{t-1})
}{m}.
\end{align}

Since $\Phi_{i,t}^2=D_{i,t}^2$, this proves
\eqref{eq:fg_noise}.
\hfill$\square$

The decomposition shows that Representation
Flux captures two effects: changes in a
neuron's average activation and random
differences between minibatches.
Therefore, high Flux does not necessarily
mean that a neuron would benefit from
splitting.

The proposition assumes independently sampled
minibatches and fixed network parameters
during evaluation. In actual FlowGenesis
training, parameters change between
minibatches, and Flux is smoothed using
an exponential moving average (EMA).
Thus, the decomposition does not exactly
describe the Flux signal used for neuron
selection.

\paragraph{Direct contribution of neuron pruning.}
Consider a neuron $i$ with final hidden-state
activation $h_i^{(S)}$ and output-weight
column $w_i$. If pruning removes only its
direct output contribution while leaving
all other hidden activations unchanged,
the output perturbation satisfies
\begin{equation}
\|f_{\mathrm{pruned}}(x)-f(x)\|
=
\|w_i\|\,|h_i^{(S)}|.
\end{equation}

This identity motivates activity-based
pruning: neurons with small activations
and bounded output weights have limited
direct influence on the output. However, removing a neuron also changes
the recurrent connectivity and may alter
the activations of remaining neurons.
Degree normalization can further modify
the effective recurrent weights.
Therefore, low activity alone does not
guarantee that pruning preserves the
network function. FlowGenesis additionally considers low
Representation Flux to identify neurons
with limited observed activation changes.
This provides a complementary heuristic
but does not guarantee that removing
such neurons is functionally harmless.

\paragraph{Proposition 3 (Local reduction of
optimization-induced activation drift
after symmetric splitting).}
Consider a neuron with output contribution
\begin{equation}
f(x;\theta)=w h(x;\theta).
\end{equation}
Suppose it is replaced by two branches
with equal output weights,
\begin{equation}
f^+(x;\theta_1,\theta_2)
=
\frac{w}{2}h(x;\theta_1)
+
\frac{w}{2}h(x;\theta_2),
\end{equation}
initialized at
$\theta_1=\theta_2=\theta$.

Let
\begin{equation}
\mu(\theta)
=
\mathbb E_x[h(x;\theta)]
\end{equation}
be differentiable, with differentiation
and expectation interchangeable.

At the splitting instant, the
optimization-induced rate of change
of each child's mean activation satisfies
\begin{equation}
\boxed{
\left.
\frac{d\mu(\theta_k)}{dt}
\right|_{\mathrm{split}}
=
\frac12
\left.
\frac{d\mu(\theta)}{dt}
\right|_{\mathrm{parent}},
\qquad k=1,2.
}
\label{eq:fg_split_flux_reduction}
\end{equation}

\emph{Proof.}
At symmetric initialization, both
branches produce the same activation
as the original neuron. By the chain
rule and equal division of the output
weight,
\begin{equation}
\nabla_{\theta_k}\mathcal L^+
=
\frac12\nabla_\theta\mathcal L.
\end{equation}

Under gradient flow,
\begin{equation}
\dot\theta_k
=
-\frac12\nabla_\theta\mathcal L,
\qquad
\dot\theta
=
-\nabla_\theta\mathcal L.
\end{equation}

Applying the chain rule to the
population mean activation gives
\begin{equation}
\frac{d\mu(\theta_k)}{dt}
=
\nabla_\theta\mu^\top\dot\theta_k
=
\frac12
\nabla_\theta\mu^\top\dot\theta.
\end{equation}

\hfill$\square$

The result shows that symmetric splitting
reduces the instantaneous optimization-induced
rate of change of each child's mean activation
by a factor of two under gradient flow.
This follows from dividing the parent's
output contribution equally between two
identical branches, thereby reducing the
gradient acting on each branch.

\subsection{Emergent Architectural Equilibrium}
\label{app:architectural_equilibrium}

FlowGenesis regulates network capacity through
Flux-guided neuron splitting and activity-dependent
pruning. We examine how these mechanisms constrain
population growth and may produce stable
architectural regimes.

\paragraph{Flux-limited population growth.}
Let $N_t$ denote the neuron population before
structural check $t$, with
\begin{equation}
N_{t+1}=N_t+B_t-D_t,
\label{eq:ae_population}
\end{equation}
where $B_t$ and $D_t$ are the numbers of
birth and pruning events.

Let $\overline{\Phi}_{i,t}\geq 0$ denote
the mean Representation Flux of neuron $i$
over the evaluation interval used to
determine birth eligibility. Splitting requires
$\overline{\Phi}_{i,t}>
\theta_{\mathrm{birth}}>0$,
together with persistence and ranking criteria.

Defining total eligibility Flux as
\begin{equation}
T_t=\sum_{i=1}^{N_t}\overline{\Phi}_{i,t},
\end{equation}
the number of neurons exceeding the threshold
is at most $T_t/\theta_{\mathrm{birth}}$.
Consequently,
\begin{equation}
\boxed{
B_t\leq
\min\left\{
r_t,\,
\frac{T_t}{\theta_{\mathrm{birth}}}
\right\},
}
\label{eq:ae_birth_bound}
\end{equation}
where $r_t$ is the per-check birth allowance.

This elementary bound connects capacity
expansion to the available Representation
Flux. However, the event limit also
constrains growth independently of Flux,
and the bound alone does not establish
population stabilization.

\paragraph{Flux--activity feedback.}
FlowGenesis prioritizes high-Flux neurons
for splitting and persistently low-activity,
low-Flux neurons for pruning.

Increasing network capacity may reduce
the prevalence of neurons eligible for
splitting while increasing the prevalence
of pruning candidates. Such behavior
would generate negative feedback on
population size, although neither
relationship is guaranteed by the
selection rules.

\paragraph{Conditional population stability.}
Consider a stationary mean-field
approximation with expected executed
birth and death counts $b(n)$ and $d(n)$.
The population update is
\begin{equation}
n_{t+1}=F(n_t)
=n_t+b(n_t)-d(n_t),
\label{eq:ae_meanfield}
\end{equation}
where $n_t$ denotes the population
in the reduced model.

An equilibrium $n^*$ satisfies
\begin{equation}
b(n^*)=d(n^*).
\label{eq:ae_equilibrium}
\end{equation}

If $b$ and $d$ are continuously
differentiable near $n^*$, the standard
sufficient condition for local
asymptotic stability of a
one-dimensional discrete-time map
\citep{elaydi2005difference} is
\begin{equation}
\boxed{
-2<
b'(n^*)-d'(n^*)
<0.
}
\label{eq:ae_local_condition}
\end{equation}

Thus, a stable mean-field equilibrium
requires sufficiently moderate negative
feedback between expected neuron birth
and pruning rates.

\paragraph{Limitations.}
These conditions characterize a reduced
population model, not the full FlowGenesis
dynamics. In particular, the required
feedback properties have not been
established for the implemented algorithm,
whose structural decisions depend on
evolving Flux, activity, eligibility
history, and population constraints.

Population stabilization may also arise
from explicit structural limits and
does not imply convergence of recurrent
connectivity or learned representations.

\subsection{Initial-Condition Forgetting in
Architectural Dynamics}
\label{app:initial_condition_forgetting}

The preceding analysis characterizes
conditions for a stable population
equilibrium. We now examine when the
influence of initial network size
decreases during architectural development.

\paragraph{Population dynamics with disturbances.}
Consider two developmental trajectories
with different initial neuron counts,
$N_0^{(a)}$ and $N_0^{(b)}$.

We represent their population updates as
\begin{equation}
N_{t+1}^{(j)}
=
\mathcal{F}(N_t^{(j)})+\xi_t^{(j)},
\qquad j\in\{a,b\},
\label{eq:ic_population}
\end{equation}
where $\mathcal{F}(n)=n+b(n)-d(n)$
is the mean-field map defined above.

The disturbance $\xi_t^{(j)}$ denotes
the deviation of realized net structural
changes from their conditional expectation
at the current population size.

\paragraph{Proposition 4 (Initial-condition
forgetting under contractive population
dynamics).}

Suppose both trajectories remain in
a neighborhood $\mathcal I$ of the
mean-field equilibrium $n^*$, where
$\mathcal{F}$ is contractive:
\begin{equation}
|\mathcal{F}(x)-\mathcal{F}(y)|
\leq q|x-y|,
\qquad
x,y\in\mathcal I,\quad 0\leq q<1.
\label{eq:ic_contraction}
\end{equation}

Assume further that the expected
disturbance discrepancy is uniformly
bounded:
\begin{equation}
\mathbb E
|\xi_t^{(a)}-\xi_t^{(b)}|
\leq\delta,
\qquad t\geq 0,
\label{eq:ic_disturbance}
\end{equation}
for some $\delta\geq 0$.

Then, for every $t\geq 0$,
\begin{equation}
\boxed{
\begin{aligned}
\mathbb E|N_t^{(a)}-N_t^{(b)}|
&\leq
q^t|N_0^{(a)}-N_0^{(b)}|\\
&\quad+
\frac{\delta(1-q^t)}{1-q}.
\end{aligned}
}
\label{eq:ic_expected_bound}
\end{equation}

\emph{Proof.}
Subtracting the population updates
and applying the triangle inequality
and contraction assumption gives
\begin{equation}
\begin{aligned}
|N_{t+1}^{(a)}-N_{t+1}^{(b)}|
&\leq
q|N_t^{(a)}-N_t^{(b)}|\\
&\quad+
|\xi_t^{(a)}-\xi_t^{(b)}|.
\end{aligned}
\label{eq:ic_recursive}
\end{equation}

Iterating this inequality yields
\begin{equation}
\begin{aligned}
|N_t^{(a)}-N_t^{(b)}|
&\leq
q^t|N_0^{(a)}-N_0^{(b)}|\\
&\quad+
\sum_{k=0}^{t-1}q^{t-1-k}
|\xi_k^{(a)}-\xi_k^{(b)}|.
\end{aligned}
\label{eq:ic_bound}
\end{equation}

Taking expectations and applying
the disturbance bound,
\begin{equation}
\begin{aligned}
\mathbb E|N_t^{(a)}-N_t^{(b)}|
&\leq
q^t|N_0^{(a)}-N_0^{(b)}|\\
&\quad+
\delta\sum_{k=0}^{t-1}q^k.
\end{aligned}
\end{equation}

Evaluating the geometric sum
establishes the result.
\hfill$\square$

Proposition 4 shows that, under
contractive population dynamics,
the contribution of initial neuron
count to the expected population
difference decreases geometrically.
Residual structural fluctuations
can nevertheless maintain differences
between developmental trajectories. For FlowGenesis, such contraction
could arise from negative feedback
between Flux-guided splitting and
activity-dependent pruning.
However, contraction is not guaranteed
by the implemented selection rules,
and the disturbance bound has not
been empirically verified. The result therefore provides a
conditional explanation for the
observed reduction in dependence
on initial network size, rather than
a guarantee of architectural convergence.
In particular, similar final neuron
counts do not imply identical recurrent
connectivity or learned representations.

\section{Architecture Visualization}
\label{app:architecture_visualization}

\begin{figure*}[p]
    \centering

    \includegraphics[width=0.83\textwidth]
        {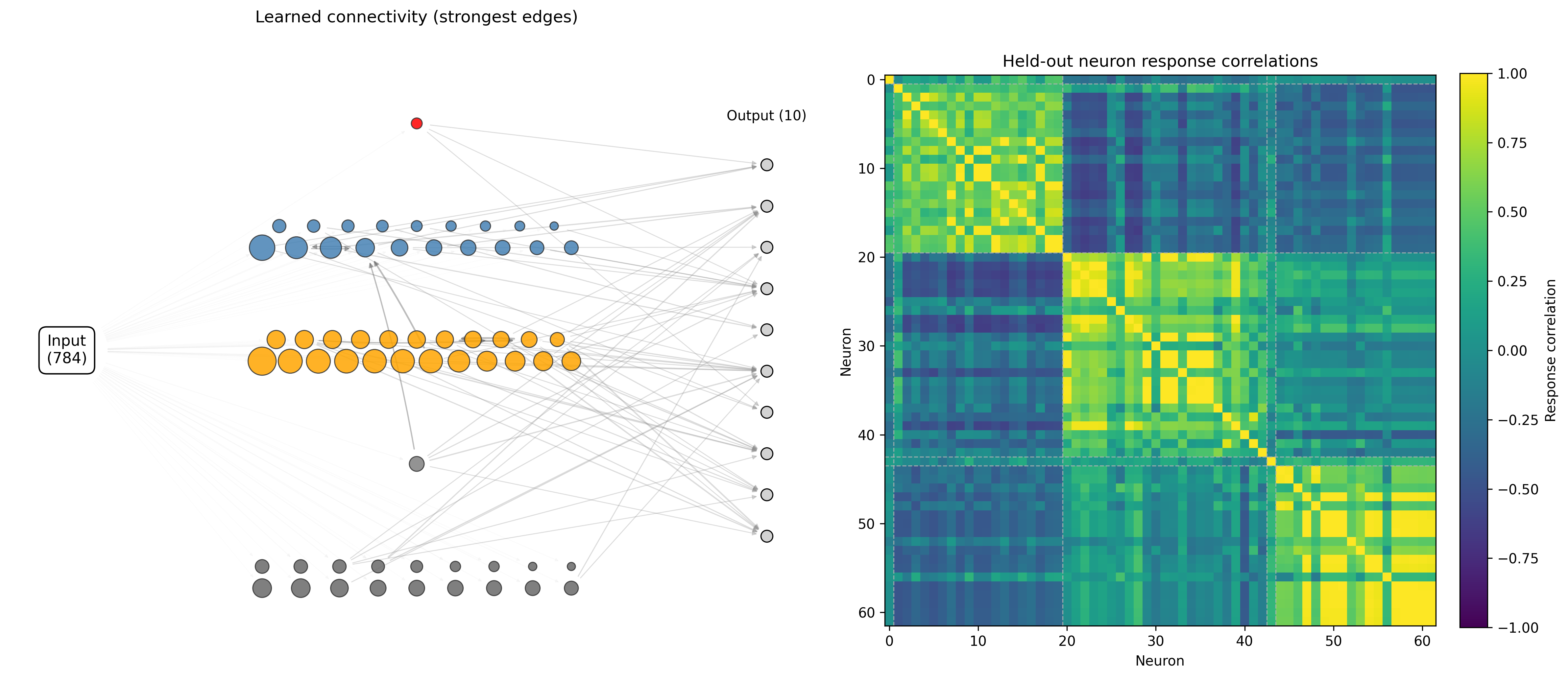}%
    \hfill
    \parbox[c]{0.10\textwidth}{\raggedright (a)}

    \vspace{0.2em}

    \includegraphics[width=0.83\textwidth]
        {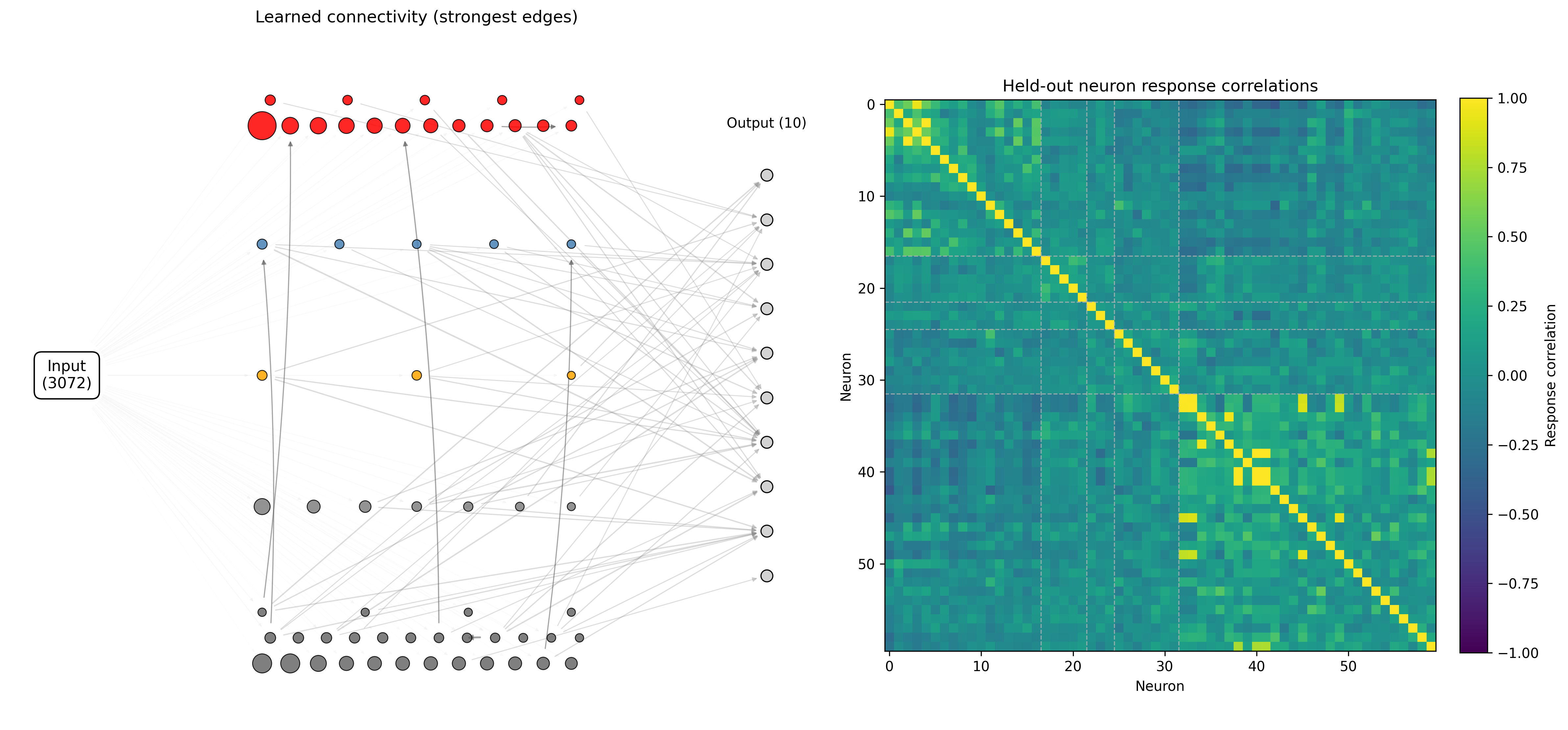}%
    \hfill
    \parbox[c]{0.10\textwidth}{\raggedright (b)}

    \vspace{0.2em}

    \includegraphics[width=0.83\textwidth]
        {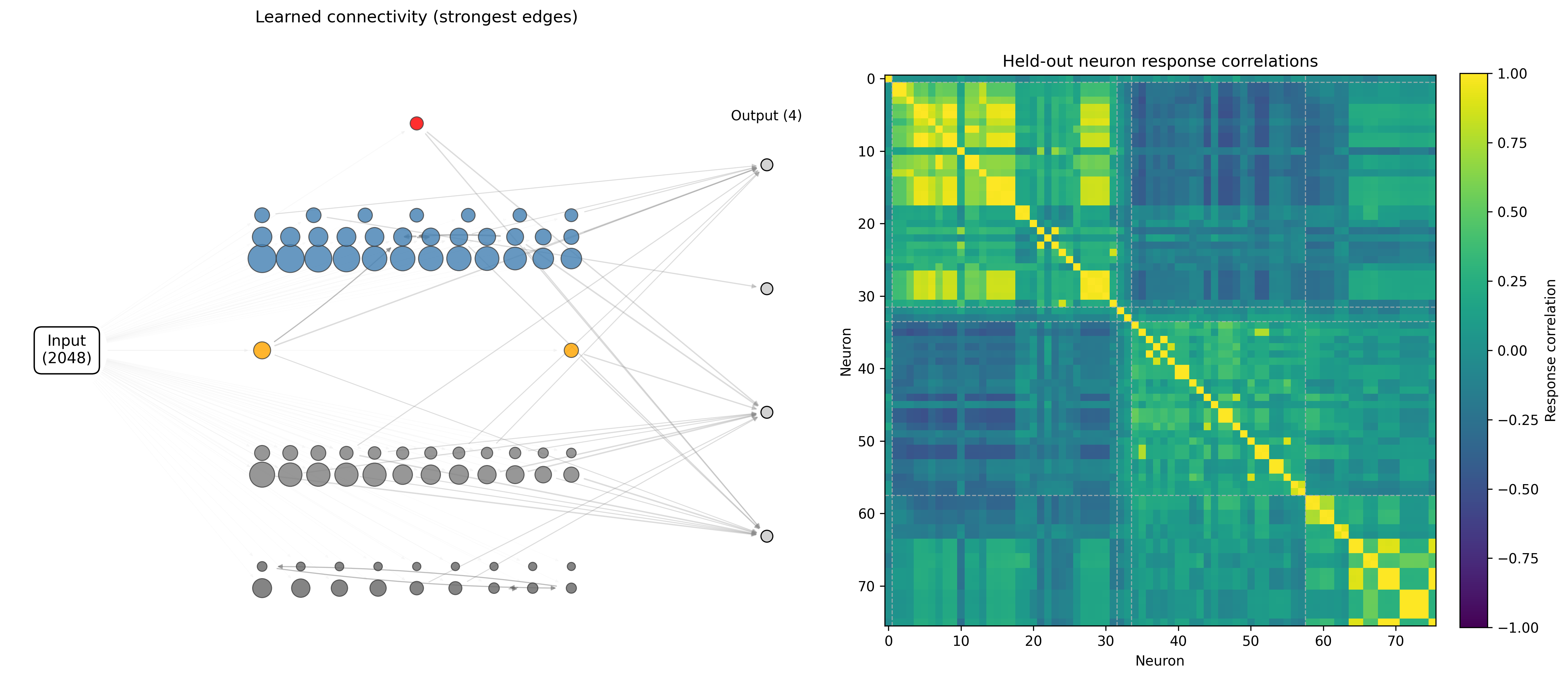}%
    \hfill
    \parbox[c]{0.10\textwidth}{\raggedright (c)}

    \caption{
        \textbf{Structural and functional organization of trained
        FlowGenesis architectures.}
        (a) Fashion-MNIST; (b) CIFAR-10; (c) AG News.
        Learned connectivity (left) and held-out neuronal response
        correlations (right). Node colors indicate correlation-based
        functional clusters; node areas reflect mean absolute activation,
        normalized within each network. Only the strongest connections
        are shown. Neurons are ordered by hierarchical clustering of
        response correlations, with dashed lines marking cluster
        boundaries. The correlation scale is fixed to $[-1,1]$
        across datasets.
    }
    \label{fig:flowgenesis_architectures}
\end{figure*}

\paragraph{Visualization methodology.}
Figure~\ref{fig:flowgenesis_architectures} presents the final
architectures learned using FlowGenesis on Fashion-MNIST,
CIFAR-10, and AG News. 
For architectural visualization, we select a
representative checkpoint using the CKA medoid
across evaluated runs.
The network is evaluated in inference mode on a randomly
sampled probe set of 1,000 examples from the test partition.
Fashion-MNIST and CIFAR-10 inputs are flattened pixel tensors,
whereas AG News uses the original cached TF-IDF representation
with the vocabulary employed during training.

To examine the functional organization of developed
networks, we record hidden-neuron activations after
recurrent relaxation for 1,000 randomly sampled test
examples using a fixed random seed. The examples are
divided equally into two disjoint subsets: one for
identifying functional clusters and the other for
evaluating their reproducibility. Using the first subset, we calculate Pearson
correlations $\rho_{ij}$ between the activation
patterns of each pair of neurons across examples.
Neurons are then grouped using average-linkage
hierarchical clustering with correlation distance
$d_{ij}=1-\rho_{ij}$, producing at most five
functional clusters. Neurons with similar activation
patterns are therefore preferentially assigned to
the same cluster. To assess whether these groups reflect consistent
functional relationships rather than correlations
specific to the clustering examples, we recompute
the correlation matrix using the second subset.
The cluster assignments remain unchanged, allowing
the discovered groups to be examined on examples
not used during clustering. For visualization,
neurons are ordered by functional cluster and,
within each cluster, by decreasing mean absolute
activation.

The left panels illustrate the learned network architectures.
Input dimensions are represented by a single aggregate
node, while output nodes correspond to classification
categories. Hidden-neuron colors indicate functional
clusters of neurons with correlated activation patterns,
while node sizes reflect mean absolute activation,
scaled independently for each network. Lines represent learned connections. Input-to-hidden
connections summarize input-weight vectors by their
$\ell_2$ norms. To reduce visual clutter, recurrent
connections are restricted to each neuron's strongest
outgoing connection by absolute effective weight
(accounting for masking and incoming-degree normalization),
with only the top 10\% of these candidates displayed.
Hidden-to-output connections are likewise restricted
to the strongest 10\% by absolute weight. The right panels show held-out Pearson correlation
matrices with the same neuron ordering. Dashed lines
mark functional cluster boundaries, and a common
color scale of $[-1,1]$ enables comparison across
networks.

\paragraph{Dataset-dependent functional organization.}
The learned architectures exhibit distinct functional
organizations. Fashion-MNIST shows three prominent
groups of strongly correlated neurons, with weaker
or negative correlations between groups. CIFAR-10
exhibits smaller, less coherent groups and generally
weaker correlations. AG News displays pronounced
correlation blocks, including a large group of
neurons with highly similar response patterns. These differences may reflect dataset characteristics.
The relatively regular visual patterns in Fashion-MNIST
may encourage overlapping neuronal responses, whereas
the greater visual diversity of CIFAR-10 may promote
more heterogeneous representations. In AG News,
sparse TF-IDF features may produce correlated responses
to shared lexical or topical patterns. Within FlowGenesis, Flux-guided neuron birth and
pruning adapt network capacity as representations
evolve. The resulting architectures exhibit
dataset-dependent functional organization, with
distinct patterns of neuronal correlations.

\section*{Code Availability}

The source code for FlowGenesis is publicly available at
\href{https://github.com/maksimkazanskii/FlowGenesis}
{\nolinkurl{github.com/maksimkazanskii/FlowGenesis}}.

\section{Use of Large Language Models}
\label{app:llm_use}

Large language models (LLMs) assisted with manuscript
polishing and code development.

The author retained full responsibility for the
research methodology, experimental design, result
verification, scientific interpretation, and final
manuscript. All LLM-assisted contributions were
reviewed and validated by the author.

\clearpage
\phantomsection
\addcontentsline{toc}{section}{Contents}

\begingroup
\linespread{1.2}\selectfont
\let\oldcontentsline\contentsline
\tableofcontents
\endgroup

\clearpage
\phantomsection
\addcontentsline{toc}{section}{References}
\bibliography{references}

@article{elsken2019nas,
  title={Neural Architecture Search: A Survey},
  author={Elsken, Thomas and Metzen, Jan Hendrik and Hutter, Frank},
  journal={Journal of Machine Learning Research},
  volume={20},
  number={55},
  pages={1--21},
  year={2019}
}

@inproceedings{liu2019darts,
  title={DARTS: Differentiable Architecture Search},
  author={Liu, Hanxiao and Simonyan, Karen and Yang, Yiming},
  booktitle={International Conference on Learning Representations},
  year={2019}
}

@inproceedings{cortes2017adanet,
  title={{AdaNet}: Adaptive Structural Learning of Artificial Neural Networks},
  author={Cortes, Corinna and Gonzalvo, Xavier and Kuznetsov, Vitaly
          and Mohri, Mehryar and Yang, Scott},
  booktitle={Proceedings of the 34th International Conference on Machine Learning},
  series={Proceedings of Machine Learning Research},
  volume={70},
  pages={874--883},
  year={2017},
  publisher={PMLR}
}

@inproceedings{chen2016net2net,
  title={{Net2Net}: Accelerating Learning via Knowledge Transfer},
  author={Chen, Tianqi and Goodfellow, Ian and Shlens, Jonathon},
  booktitle={International Conference on Learning Representations},
  year={2016}
}

@inproceedings{wei2016networkmorphism,
  title={Network Morphism},
  author={Wei, Tao and Wang, Changhu and Rui, Yong and Chen, Chang Wen},
  booktitle={Proceedings of the 33rd International Conference on Machine Learning},
  series={Proceedings of Machine Learning Research},
  volume={48},
  pages={564--572},
  year={2016},
  publisher={PMLR}
}

@inproceedings{wu2019splitting,
  title={Splitting Steepest Descent for Growing Neural Architectures},
  author={Wu, Lemeng and Wang, Dilin and Liu, Qiang},
  booktitle={Advances in Neural Information Processing Systems},
  volume={32},
  year={2019}
}

@inproceedings{wu2020firefly,
  title={Firefly Neural Architecture Descent: A General Approach for Growing Neural Networks},
  author={Wu, Lemeng and Liu, Bo and Stone, Peter and Liu, Qiang},
  booktitle={Advances in Neural Information Processing Systems},
  volume={33},
  year={2020}
}

@inproceedings{maile2022neurons,
  title     = {When, Where, and How to Add New Neurons to {ANNs}},
  author    = {Maile, Kaitlin and Rachelson, Emmanuel
               and Luga, Herv{\'e} and Wilson, Dennis George},
  booktitle = {Proceedings of the First International Conference
               on Automated Machine Learning},
  series    = {Proceedings of Machine Learning Research},
  volume    = {188},
  pages     = {18/1--12},
  year      = {2022},
  publisher = {PMLR},
  url       = {https://proceedings.mlr.press/v188/maile22a.html}
}

@inproceedings{evci2022gradmax,
  title     = {{GradMax}: Growing Neural Networks
               using Gradient Information},
  author    = {Evci, Utku and van Merri{\"e}nboer, Bart
               and Unterthiner, Thomas and Vladymyrov, Max
               and Pedregosa, Fabian},
  booktitle = {International Conference on Learning Representations},
  year      = {2022},
  url       = {https://openreview.net/forum?id=qjN4h_wwUO}
}

@article{dai2019nest,
  title   = {{NeST}: A Neural Network Synthesis Tool
             Based on a Grow-and-Prune Paradigm},
  author  = {Dai, Xiaoliang and Yin, Hongxu
             and Jha, Niraj K.},
  journal = {IEEE Transactions on Computers},
  volume  = {68},
  number  = {10},
  pages   = {1487--1497},
  year    = {2019},
  doi     = {10.1109/TC.2019.2914438}
}

@inproceedings{evci2020rigl,
  title={Rigging the Lottery: Making All Tickets Winners},
  author={Evci, Utku and Gale, Trevor and Menick, Jacob
          and Castro, Pablo Samuel and Elsen, Erich},
  booktitle={Proceedings of the 37th International Conference on Machine Learning},
  series={Proceedings of Machine Learning Research},
  volume={119},
  pages={2943--2952},
  year={2020},
  publisher={PMLR}
}

@inproceedings{yoon2018den,
  title={Lifelong Learning with Dynamically Expandable Networks},
  author={Yoon, Jaehong and Yang, Eunho and Lee, Jeongtae and Hwang, Sung Ju},
  booktitle={International Conference on Learning Representations},
  year={2018}
}

@inproceedings{pashakhanloo2023drift,
  title={Stochastic Gradient Descent-Induced Drift of Representation in a Two-Layer Neural Network},
  author={Pashakhanloo, Farhad and Koulakov, Alexei},
  booktitle={Proceedings of the 40th International Conference on Machine Learning},
  series={Proceedings of Machine Learning Research},
  volume={202},
  pages={27401--27419},
  year={2023},
  publisher={PMLR}
}

@inproceedings{lyle2022capacity,
  title     = {Understanding and Preventing Capacity Loss
               in Reinforcement Learning},
  author    = {Lyle, Clare and Rowland, Mark and Dabney, Will},
  booktitle = {International Conference on Learning Representations},
  year      = {2022},
  url       = {https://openreview.net/forum?id=ZkC8wKoLbQ7}
}

@inproceedings{plantec2024lndp,
  title     = {Evolving Self-Assembling Neural Networks:
               From Spontaneous Activity to
               Experience-Dependent Learning},
  author    = {Plantec, Erwan and Pedersen, Joachim W.
               and Montero, Milton L. and Nisioti, Eleni
               and Risi, Sebastian},
  booktitle = {ALIFE 2024: Proceedings of the
               2024 Artificial Life Conference},
  pages     = {37},
  publisher = {MIT Press},
  year      = {2024},
  doi       = {10.1162/isal_a_00755}
}

@inproceedings{najarro2023ndp,
  title={Towards Self-Assembling Artificial Neural Networks through
         Neural Developmental Programs},
  author={Najarro, Elias and Sudhakaran, Shyam and Risi, Sebastian},
  booktitle={Artificial Life Conference Proceedings},
  year={2023},
  doi={10.1162/isal_a_00697}
}

@article{jia2026smgrnn,
  title   = {Self-Motivated Growing Neural Network
             for Adaptive Architecture via
             Local Structural Plasticity},
  author  = {Jia, Yiyang and Zhou, Chengxu},
  journal = {Neurocomputing},
  volume  = {702},
  pages   = {134581},
  year    = {2026},
  doi     = {10.1016/j.neucom.2026.134581},
  url     = {https://doi.org/10.1016/j.neucom.2026.134581}
}

@inproceedings{ferigo2023selfbuilding,
  title={Self-Building Neural Networks},
  author={Ferigo, Andrea and Iacca, Giovanni},
  booktitle={Companion Conference on Genetic and Evolutionary Computation},
  pages={643--646},
  year={2023},
  publisher={ACM},
  doi={10.1145/3583133.3590531}
}

@article{dohare2021continual,
  title   = {{Continual Backprop}: Stochastic Gradient Descent
             with Persistent Randomness},
  author  = {Dohare, Shibhansh and Sutton, Richard S.
             and Mahmood, A. Rupam},
  journal = {arXiv preprint arXiv:2108.06325},
  year    = {2021},
  doi     = {10.48550/arXiv.2108.06325},
  url     = {https://arxiv.org/abs/2108.06325}
}

@article{holtmaat2009experience,
  title={Experience-dependent structural synaptic plasticity in the
         mammalian brain},
  author={Holtmaat, Anthony and Svoboda, Karel},
  journal={Nature Reviews Neuroscience},
  volume={10},
  number={9},
  pages={647--658},
  year={2009}
}

@article{paolicelli2011synaptic,
  title={Synaptic pruning by microglia is necessary for normal brain
         development},
  author={Paolicelli, Rosa C. and Bolasco, Giuseppina and Pagani, Luca
          and Maggi, Laura and Scianni, Marta and Panzanelli, Patrizia
          and Giustetto, Maurizio and Ferreira, Tiago A. and Guiducci,
          Elisa and Dumas, Ludovic and Ragozzino, Davide and Gross,
          Cornelius T.},
  journal={Science},
  volume={333},
  number={6048},
  pages={1456--1458},
  year={2011}
}

@article{ming2011adult,
  title={Adult Neurogenesis in the Mammalian Brain: Significant Answers
         and Significant Questions},
  author={Ming, Guo-li and Song, Hongjun},
  journal={Neuron},
  volume={70},
  number={4},
  pages={687--702},
  year={2011}
}

@techreport{jaeger2001echo,
  author      = {Jaeger, Herbert},
  title       = {The ``Echo State'' Approach to Analysing and Training Recurrent Neural Networks},
  institution = {German National Research Center for Information Technology},
  year        = {2001},
  number      = {GMD Report 148}
}

@article{maass2002realtime,
  author  = {Maass, Wolfgang and Natschl{\"a}ger, Thomas and Markram, Henry},
  title   = {Real-Time Computing Without Stable States: A New Framework for Neural Computation Based on Perturbations},
  journal = {Neural Computation},
  volume  = {14},
  number  = {11},
  pages   = {2531--2560},
  year    = {2002},
  doi     = {10.1162/089976602760407955}
}

@article{welch1951comparison,
  author  = {Welch, B. L.},
  title   = {On the Comparison of Several Mean Values: An Alternative Approach},
  journal = {Biometrika},
  volume  = {38},
  number  = {3--4},
  pages   = {330--336},
  year    = {1951},
  doi     = {10.1093/biomet/38.3-4.330}
}

@article{holm1979simple,
  author  = {Holm, Sture},
  title   = {A Simple Sequentially Rejective Multiple Test Procedure},
  journal = {Scandinavian Journal of Statistics},
  volume  = {6},
  number  = {2},
  pages   = {65--70},
  year    = {1979}
}

@article{verstraeten2007experimental,
  author  = {Verstraeten, David and Schrauwen, Benjamin and D'Haene, Michiel and Stroobandt, Dirk},
  title   = {An Experimental Unification of Reservoir Computing Methods},
  journal = {Neural Networks},
  volume  = {20},
  number  = {3},
  pages   = {391--403},
  year    = {2007},
  doi     = {10.1016/j.neunet.2007.04.003}
}

@inproceedings{kornblith2019similarity,
  author    = {Kornblith, Simon and Norouzi, Mohammad and Lee, Honglak and Hinton, Geoffrey},
  title     = {Similarity of Neural Network Representations Revisited},
  booktitle = {Proceedings of the 36th International Conference on Machine Learning},
  series    = {Proceedings of Machine Learning Research},
  volume    = {97},
  pages     = {3519--3529},
  year      = {2019},
  publisher = {PMLR}
}

@article{kuhn1955hungarian,
  author  = {Kuhn, Harold W.},
  title   = {The Hungarian Method for the Assignment Problem},
  journal = {Naval Research Logistics Quarterly},
  volume  = {2},
  number  = {1--2},
  pages   = {83--97},
  year    = {1955},
  doi     = {10.1002/nav.3800020109}
}

@techreport{krizhevsky2009learning,
  author      = {Krizhevsky, Alex},
  title       = {Learning Multiple Layers of Features from Tiny Images},
  institution = {University of Toronto},
  year        = {2009}
}

@misc{kazanskii2026geometry,
  title={Geometry of Forgetting: Representation Flux in Continual Learning},
  author={Kazanskii, Maksim A.},
  year={2026},
  eprint={2608.15854},
  archivePrefix={arXiv},
  primaryClass={cs.LG},
  doi={10.48550/arXiv.2608.15854}
}

@article{turrigiano2004homeostatic,
  title={Homeostatic plasticity in the developing nervous system},
  author={Turrigiano, Gina G. and Nelson, Sacha B.},
  journal={Nature Reviews Neuroscience},
  volume={5},
  pages={97--107},
  year={2004},
  doi={10.1038/nrn1327}
}

@article{turrigiano2012homeostatic,
  title={Homeostatic synaptic plasticity: local and global mechanisms for stabilizing neuronal function},
  author={Turrigiano, Gina},
  journal={Cold Spring Harbor Perspectives in Biology},
  volume={4},
  number={1},
  pages={a005736},
  year={2012},
  doi={10.1101/cshperspect.a005736}
}

@article{butz2013structural,
  title={A Simple Rule for Dendritic Spine and Axonal Bouton Formation Can Account for Cortical Reorganization after Focal Retinal Lesions},
  author={Butz, Markus and van Ooyen, Arjen},
  journal={PLOS Computational Biology},
  volume={9},
  number={10},
  pages={e1003259},
  year={2013},
  doi={10.1371/journal.pcbi.1003259}
}

@article{butz2014structural,
  title={Homeostatic structural plasticity can account for topology changes following deafferentation and focal stroke},
  author={Butz, Markus and Steenbuck, Ines D. and van Ooyen, Arjen},
  journal={Frontiers in Neuroanatomy},
  volume={8},
  pages={115},
  year={2014},
  doi={10.3389/fnana.2014.00115}
}

@inproceedings{fritzke1995growing,
  title     = {A Growing Neural Gas Network Learns Topologies},
  author    = {Fritzke, Bernd},
  booktitle = {Advances in Neural Information Processing Systems 7},
  pages     = {625--632},
  publisher = {MIT Press},
  year      = {1995},
  url       = {https://papers.nips.cc/paper/1994/hash/d56b9fc4b0f1be8871f5e1c40c0067e7-Abstract.html}
}

@article{gross2008adaptive,
  title={Adaptive Coevolutionary Networks: A Review},
  author={Gross, Thilo and Blasius, Bernd},
  journal={Journal of the Royal Society Interface},
  volume={5},
  number={20},
  pages={259--271},
  year={2008},
  doi={10.1098/rsif.2007.1229}
}

@article{rusu2016progressive,
  title={Progressive Neural Networks},
  author={Rusu, Andrei A. and Rabinowitz, Neil C.
          and Desjardins, Guillaume and Soyer, Hubert
          and Kirkpatrick, James and Kavukcuoglu, Koray
          and Pascanu, Razvan and Hadsell, Raia},
  journal={arXiv preprint arXiv:1606.04671},
  year={2016},
  url={https://arxiv.org/abs/1606.04671}
}

@article{martinetz1993neural,
  title   = {Neural-Gas Network for Vector Quantization and Its Application to Time-Series Prediction},
  author  = {Martinetz, Thomas M. and Berkovich, Stanislav G. and Schulten, Klaus J.},
  journal = {IEEE Transactions on Neural Networks},
  volume  = {4},
  number  = {4},
  pages   = {558--569},
  year    = {1993},
  doi     = {10.1109/72.238311}
}

@inproceedings{zhang2015character,
  title     = {Character-level Convolutional Networks
               for Text Classification},
  author    = {Zhang, Xiang and Zhao, Junbo
               and LeCun, Yann},
  booktitle = {Advances in Neural Information
               Processing Systems},
  volume    = {28},
  year      = {2015}
}

@article{tashiro2006nmda,
  title={NMDA-receptor-mediated, cell-specific
         integration of new neurons in adult
         dentate gyrus},
  author={Tashiro, Ayumu and Sandler, Vladislav M.
          and Toni, Nicolas and Zhao, Chunmei
          and Gage, Fred H.},
  journal={Nature},
  volume={442},
  pages={929--933},
  year={2006},
  doi={10.1038/nature05028}
}

@article{schafer2012microglia,
  title={Microglia Sculpt Postnatal Neural Circuits
         in an Activity and Complement-Dependent Manner},
  author={Schafer, Dorothy P. and Lehrman, Emily K.
          and Kautzman, Amanda G. and others},
  journal={Neuron},
  volume={74},
  number={4},
  pages={691--705},
  year={2012},
  doi={10.1016/j.neuron.2012.03.026}
}

@article{xiao2017fashion,
  title={Fashion-MNIST: a Novel Image Dataset for Benchmarking Machine Learning Algorithms},
  author={Xiao, Han and Rasul, Kashif and Vollgraf, Roland},
  journal={arXiv preprint arXiv:1708.07747},
  year={2017}
}

@article{lecun1998gradient,
  title={Gradient-Based Learning Applied to Document Recognition},
  author={LeCun, Yann and Bottou, Leon and Bengio, Yoshua and Haffner, Patrick},
  journal={Proceedings of the IEEE},
  volume={86},
  number={11},
  pages={2278--2324},
  year={1998}
}

@book{elaydi2005difference,
  author    = {Elaydi, Saber},
  title     = {An Introduction to Difference Equations},
  edition   = {3},
  series    = {Undergraduate Texts in Mathematics},
  publisher = {Springer},
  address   = {New York},
  year      = {2005},
  doi       = {10.1007/0-387-27602-5}
}

\end{document}